\documentclass{article}

\PassOptionsToPackage{numbers}{natbib}
    \usepackage[preprint]{neurips_data_2024}

\usepackage[utf8]{inputenc} % allow utf-8 input
\usepackage[T1]{fontenc}    % use 8-bit T1 fonts
\usepackage{hyperref}       % hyperlinks
\usepackage{url}            % simple URL typesetting
\usepackage{booktabs}       % professional-quality tables
\usepackage{amsfonts}       % blackboard math symbols
\usepackage{nicefrac}       % compact symbols for 1/2, etc.
\usepackage{microtype}      % microtypography
\usepackage{xcolor}         % colors
\usepackage{colortbl}
\usepackage{makecell}
\usepackage{graphicx}
\usepackage{subfigure}
\usepackage{enumitem}   % enumerate
\usepackage{multirow}   % for table
\usepackage{float}
\usepackage{wrapfig}

\title{FedLLM-Bench: Realistic Benchmarks for \\ Federated Learning of Large Language Models}

\author{%
  Rui Ye\textsuperscript{1}\footnotemark[1] $\quad$ Rui Ge\textsuperscript{1}\footnotemark[1] $\quad$ Xinyu Zhu\textsuperscript{1} $\quad$
  \textbf{Jingyi Chai}\textsuperscript{1}  $\quad$ \textbf{Yaxin Du}\textsuperscript{1} \\ 
  \textbf{Yang Liu\textsuperscript{2}} $\quad$
  \textbf{Yanfeng Wang\textsuperscript{3,1}} $\quad$ \textbf{Siheng Chen\textsuperscript{1,3}} \\\\
  \textsuperscript{1} Shanghai Jiao Tong University $\quad$ \textsuperscript{2} Tsinghua University $\quad$ \textsuperscript{3} Shanghai AI Laboratory \\
}

\begin{document}

\maketitle
\renewcommand{\thefootnote}{\fnsymbol{footnote}}
\footnotetext[1]{Equal contribution. 
% \qquad \footnotemark[2]Corresponding author.
}

\begin{abstract}

Federated learning has enabled multiple parties to collaboratively train large language models without directly sharing their data (FedLLM).
Following this training paradigm, the community has put massive efforts from diverse aspects including framework, performance, and privacy.
However, an unpleasant fact is that there are currently no realistic datasets and benchmarks for FedLLM and previous works all rely on artificially constructed datasets, failing to capture properties in real-world scenarios.
Addressing this, we propose FedLLM-Bench, which involves 8 training methods, 4 training datasets, and 6 evaluation metrics, to offer a comprehensive testbed for the FedLLM community.
FedLLM-Bench encompasses three datasets (e.g., user-annotated multilingual dataset) for federated instruction tuning and one dataset (e.g., user-annotated preference dataset) for federated preference alignment, whose scale of client number ranges from 38 to 747.
Our datasets incorporate several representative diversities: language, quality, quantity, instruction, length, embedding, and preference, capturing properties in real-world scenarios.
Based on FedLLM-Bench, we conduct experiments on all datasets to benchmark existing FL methods and provide empirical insights (e.g., multilingual collaboration).
We believe that our FedLLM-Bench can benefit the FedLLM community by reducing required efforts, providing a practical testbed, and promoting fair comparisons.
Code and datasets are available at \href{https://github.com/rui-ye/FedLLM-Bench}{https://github.com/rui-ye/FedLLM-Bench}.
\end{abstract}

\section{Introduction}
\label{sec:intro}
Large language models (LLMs) have achieved unprecedented success in diverse domains~\cite{openai2023gpt4,ouyang2022training,wei2022finetuned,chen2024agentverse,hong2024metagpt,jablonka2024leveraging}.
These LLMs are usually trained by centralized learning paradigm, where various parties individually collect massive data for model training.
In this case, the data amount of each individual party is hard to scale due to the high cost of collecting and annotating data.
However, their data cannot be directly shared for collaboration due to property and privacy issues.

To relieve the required cost of each party, federated learning (FL)~\cite{fedavg,kairouz2021advances} has emerged as a sound and off-the-shelf technique to facilitate collaboration, which leverages decentralized language data to collaboratively train LLMs in a privacy-preserving way (FedLLM)~\cite{ye2024openfedllm,federatedscopellm,fedml-fedllm}.
To facilitate the development of FedLLM, there has been a series of code frameworks such as OpenFedLLM~\cite{ye2024openfedllm}, FederatedScope-LLM~\cite{federatedscopellm}, and FedML-LLM~\cite{fedml-fedllm}; and many methods that tackle the issues of data quality~\cite{zhao2024enhancing}, intellectual property protection~\cite{wu2024fedbiot}, privacy~\cite{sun2024improving}, limited resources~\cite{fang2024automated} in FedLLM.

Despite that massive efforts have been made, one significant concern remains: there is currently no realistic benchmark for FedLLM, making it hard to practically evaluate the effectiveness of FL methods in real-world scenarios.
In such context, each previous work constructs its own FL datasets by artificially partitioning existing centralized datasets~\cite{ye2024openfedllm,federatedscopellm,wu2024fedbiot}, falling short of capturing the natural properties existed in real-world cross-user datasets~\cite{ogier2022flamby,song2022flair}.
Even worse, these papers often follow different training and evaluation setups, which significantly increases the difficulty of re-implementations and risk of unfair comparisons~\cite{zhang2024fedpit,sun2024improving}.

To fill this gap, we propose the first realistic benchmark for FedLLM termed FedLLM-Bench, offering a comprehensive testbed for the FedLLM community.
FedLLM-Bench encompasses three datasets for federated instruction tuning (including one user-annotated multilingual dataset: Fed-Aya, and two datasets with realistic user instructions: Fed-WildChat and Fed-ChatbotIT) and one dataset (user-annotated preference dataset: Fed-ChatbotPA) for federated preference alignment.
These datasets are all naturally split by real-world user ID with the scale ranging from 38 to 747 clients, therefore exhibiting realistic federated properties (especially for cross-device setup in FL where data are partitioned by user devices).
Specifically, datasets in our FedLLM-Bench inherit the following diversities (Table~\ref{tab:dataset_info}):
\textbf{(1) Language:} clients' datasets (e.g., our Fed-Aya dataset) cover data from diverse languages, modeling the real-world scenarios of multilingual collaboration.
\textbf{(2) Quality and Quantity:} the quality and quantity of clients' datasets vary across each other, which is a common property in real-world applications.
\textbf{(3) Length:} the sequence length of clients' data could be quite different, representing a new type of data heterogeneity in FL.
\textbf{(4) Preference:} different clients have different preferences as verified by different preferred instructions in instruction tuning datasets (e.g., Fed-WildChat) and different preferred responses in preference alignment dataset (i.e., Fed-ChatbotPA), mirroring the complexities of real-world data scenarios.
These diversities make our FedLLM-Bench a comprehensive benchmark in the era of FedLLM, serving as a great successor to representative benchmarks for classical tasks such as LEAF~\cite{caldas2018leaf} benchmark.

Based on these datasets, we implement 8 representative baseline methods and 6 evaluation metrics, and conduct extensive experiments.
Our experiments mainly demonstrate (1) that federated learning can consistently bring performance gain compared to local training without collaboration; and (2) the performance ranking of several representative baseline methods.
Besides serving as a benchmark for performance comparison, our FedLLM-Bench can also support exploration of new research directions thanks to its flexibility and diversity.
As an example, we conduct an exploratory experiment based on the multilingual dataset Fed-Aya, showing that collaboration among similar languages could potentially bring more benefits comparing to collaboration among all languages.

Our contributions are as follows:
\begin{enumerate}[leftmargin=*]
    \item We propose the first realistic FedLLM benchmark, FedLLM-Bench, which encompasses four naturally split datasets. FedLLM-Bench covers diverse tasks, scales, languages, qualities, quantities, lengths, and preferences, mirroring the complexities and diversities of real-world scenarios.
    \item We integrate these datasets into a codebase with 8 representative baseline methods and 6 evaluation metrics, and open-source the datasets with the integrated codebase for the community.
    \item We conduct extensive experiments to demonstrate the status of several existing baseline methods on our FedLLM-Bench and show its potential in promoting exploration of new research directions.
\end{enumerate}

\section{Related work}

\textbf{Federated learning for large language models.}
Federated learning is a privacy-preserving and collaborative training paradigm that enables multiple parties to collaboratively train a shared global model without sharing their raw data~\cite{fedavg,kairouz2021advances,han2022fedx,park2023feddefender,mai2023vertical,yi2021efficient}.
Data heterogeneity is one of the most representative challenges in FL, where clients' datasets are drawn from different distributions~\cite{fedavgm,ye2023feddisco,li2023no,fan2024locally,mai2024rflpa,fan2024federated}.
Addressing this, numerous methods have been proposed by regularization~\cite{fedprox}, gradient correction~\cite{scaffold}, feature alignment~\cite{moon,zhang2024upload,zhang2024fedtgp}, adjustment of aggregation weights~\cite{ye2023feddisco,fednova,li2023revisiting}, introducing momentum~\cite{fedavgm,fedopt}, or leveraging pre-trained models~\cite{nguyen2023begin,chen2023importance}.

Recently, having witnessed the success of large language models (LLMs) in centralized learning~\cite{openai2023gpt4,llama,llama2,wei2022finetuned,jiang2023mistral}, many researchers start to explore training LLMs via federated learning, mitigating the issue of the shortage of public data or private data of one individual~\cite{ouyang2022training,wu2023bloomberggpt,xue2023db,ye2024openfedllm}.
Within one year, there have been many frameworks such as OpenFedLLM~\cite{ye2024openfedllm}, FederatedScope-LLM~\cite{federatedscopellm}, FedML-LLM~\cite{fedml-fedllm}, and diverse methods such as FedbiOT~\cite{wu2024fedbiot} that protects model property and FFA-LoRA~\cite{sun2024improving} that improves performance under differential privacy.

However, one significant issue of these previous works is that their experiments are all based on artificially crafted FL datasets, falling short of extrapolating their effectiveness in real-world scenarios.
Addressing this, we propose the first realistic benchmark for FedLLM, FedLLM-Bench, which mirrors the complexities and diversities in real-world applications.
Besides, we implement 8 representative baseline methods in our FedLLM-Bench to demonstrate their effectiveness in realistic scenarios.

\begin{table}[t]
    \centering
    \setlength\tabcolsep{5pt}
    \small
    \caption{Summary of our four realistic FedLLM datasets. IT denotes instruction tuning and PA denotes preference alignment. \# denotes `the number of' and L. denotes `the length of'. Our datasets exhibit diversities in characteristic, task, client number, quantity, length, and quality.}
    \label{tab:dataset_info}
    \begin{tabular}{l|cccc}
    \toprule
     \rowcolor{blue!5}   Dataset Name & Fed-Aya~\cite{singh2024aya} & Fed-ChatbotIT~\cite{chatbot} & Fed-WildChat~\cite{zhao2024wildchat} & Fed-ChatbotPA~\cite{chatbot} \\
    \midrule
        \rowcolor{blue!8} Characteristic & Multilingual & Single-Turn chat & Multi-Turn chat & Preference\\
        \rowcolor{blue!8} Applied Task & IT & IT & IT & PA  \\ 
        \rowcolor{gray!10} \# Clients (Total) & 38 & 237 & 100  & 747 \\
        \rowcolor{gray!10} \# Samples (Total) & 25,513 & 6,166 & 52,703 & 9,508  \\
        \rowcolor{gray!15} \# Samples (Client) & 671 $\pm$ 815 & 26 $\pm$ 33 & 527 $\pm$ 477 & 13 $\pm$ 21 \\
        \rowcolor{gray!15} L. Instruction (Client) & 116 $\pm$ 199  & 68 $\pm$ 119 & 331 $\pm$ 435  & 69 $\pm$ 124 \\
        \rowcolor{gray!15}  L. Response (Client) & 225 $\pm$ 411 & 211 $\pm$ 176 & 506 $\pm$ 470 & 218 $\pm$ 178 \\
        \rowcolor{gray!15} Data Quality (Client) & 0.63 $\pm$ 0.28 & 0.67 $\pm$ 0.22 & 0.79 $\pm$ 0.37 & 0.68 $\pm$ 0.21 \\
    \bottomrule
    \end{tabular}
\end{table}

\textbf{Datasets and benchmarks in federated learning.}
Since clients' data are collected independently without coordination, the issue of data heterogeneity commonly exists in FL.
A large proportion of FL works~\cite{fedavg,ye2023feddisco,fedprox,moon} simulate data heterogeneity by artificially partitioning classic datasets such as CIFAR-10/100~\cite{cifar10}, Fashion-MNIST~\cite{xiao2017fashion}, and MNIST~\cite{deng2012mnist}.
Addressing this, several realistic benchmarks are proposed for classic tasks such as image and text classification, which include LEAF~\cite{caldas2018leaf} (a suite of user-split datasets), FLAIR~\cite{song2022flair} (multi-label image classification), and FLamby~\cite{ogier2022flamby} (a benchmark for medical analysis).
However, currently, there is no realistic dataset or benchmark for the tasks of FedLLM, while our FedLLM-Bench stands out as the first one in the literature.
Besides, our FedLLM-Bench covers two unique tasks compared to previous benchmarks: federated instruction tuning~\cite{longpre2023flan,xu2023wizardlm,ye2024openfedllm} and federated preference alignment~\cite{kirk2023past,rafailov2024direct,ye2024openfedllm}.

\section{FedLLM-Bench: a realistic benchmark for FedLLM}

Here, we introduce our FedLLM-Bench, from four perspectives: training methods, descriptions of training datasets, analysis of training datasets, and evaluation metrics.

\subsection{Training methods}

\textbf{FedLLM overview.}
FedLLM involves four iterative steps: server-to-client model downloading, local model training, client-to-server model uploading, and global model aggregation.
During FedLLM, clients could collaborate on two critical tasks for LLMs~\cite{ouyang2022training}: instruction tuning and preference alignment, which are challenging for individuals due to high cost of data collection~\cite{zhou2023lima,ji2024beavertails}.
Besides, various FL baseline methods~\cite{fedprox,scaffold,fedopt} can be incorporated into FedLLM.

\textbf{Tasks: instruction tuning \& preference alignment.}
In instruction tuning~\cite{ouyang2022training,ye2024openfedllm}, each data sample is an instruction-response pair, where the LLMs are trained to follow instructions to generate the expected responses via supervised fine-tuning.
In preference alignment~\cite{bai2022constitutional,ye2024openfedllm}, each data sample consists of an instruction, a preferred and a dispreferred response, where the LLMs are trained to align with the preferred response given user instructions via direct preference optimization~\cite{rafailov2024direct}.
For both two tasks, we adopt the most commonly used parameter-efficient fine-tuning technique LoRA~\cite{hu2021lora}, reducing the requirements of computation and communication in FL~\cite{yang_survey,wangsurvey}.

\textbf{FL: baseline methods.}
In our FedLLM-Bench, we implement 8 representative baseline methods, including local training without collaboration and 7 classical FL baseline methods.
Following the standard baseline FedAvg~\cite{fedavg}, at the local training part, we implement FedProx~\cite{fedprox} which applies local-global model regularization and SCAFFOLD~\cite{scaffold} which introduces control variate to correct local gradients; while at the model aggregation part, we implement FedAvgM~\cite{fedavgm}, FedAdagrad~\cite{fedopt}, FedYogi~\cite{fedopt}, and FedAdam~\cite{fedopt} which introduce model momentum to update global model.

\subsection{Descriptions of training datasets}

\begin{figure}[t]
    \centering
    \subfigure[Language (Fed-Aya)]{ 
    \centering
    \includegraphics[width=0.21\linewidth]{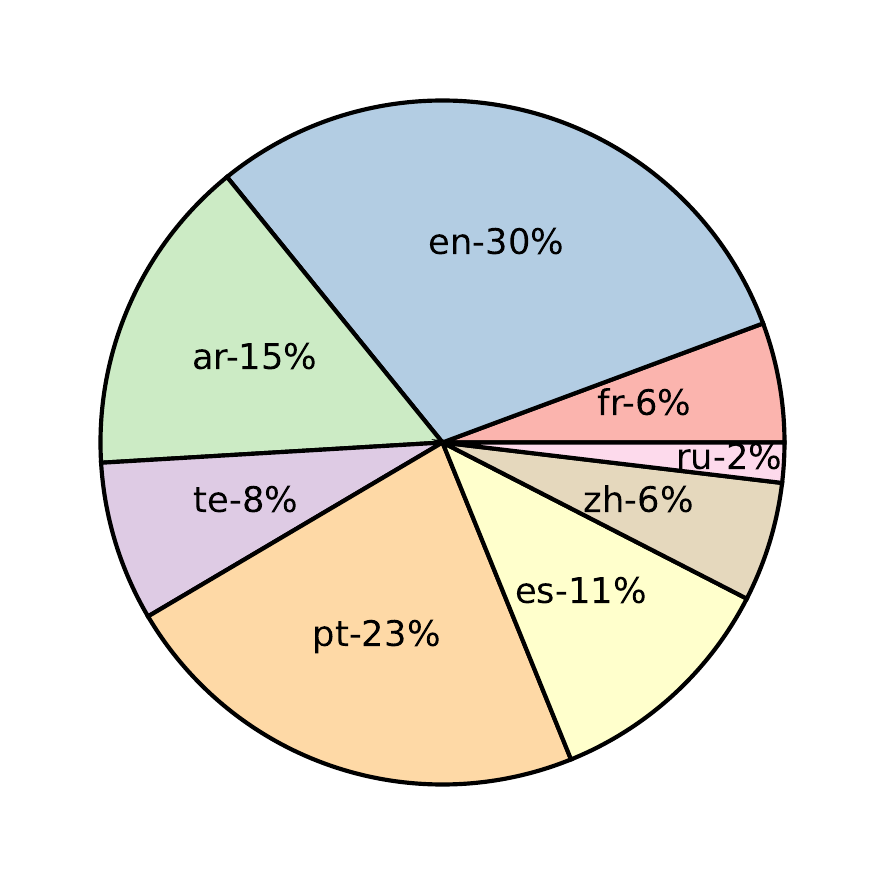}
    \label{fig:aya_pie}
    }
    \subfigure[Length Distribution]{   
    \centering    
    \includegraphics[width=0.2375\linewidth]{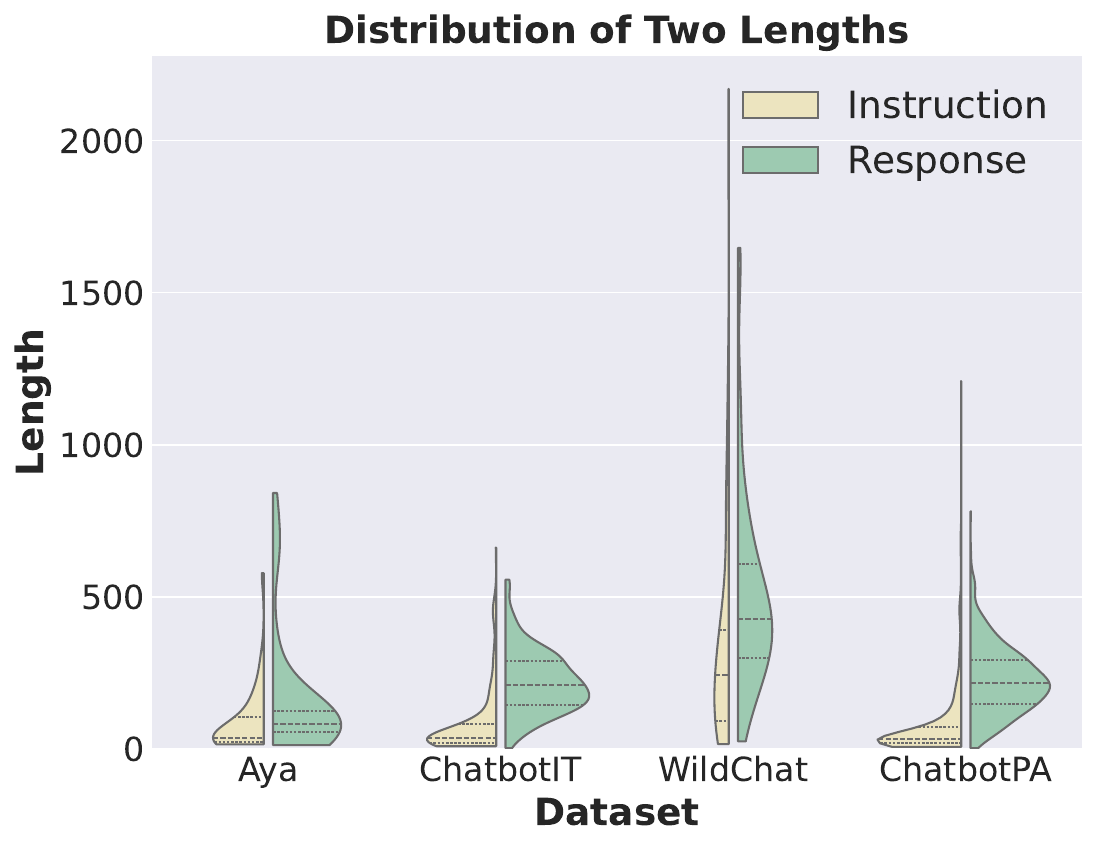}
    \label{fig:length_violin}
    }
    \subfigure[Length Preference]{ 
    \centering
    \includegraphics[width=0.235\linewidth]{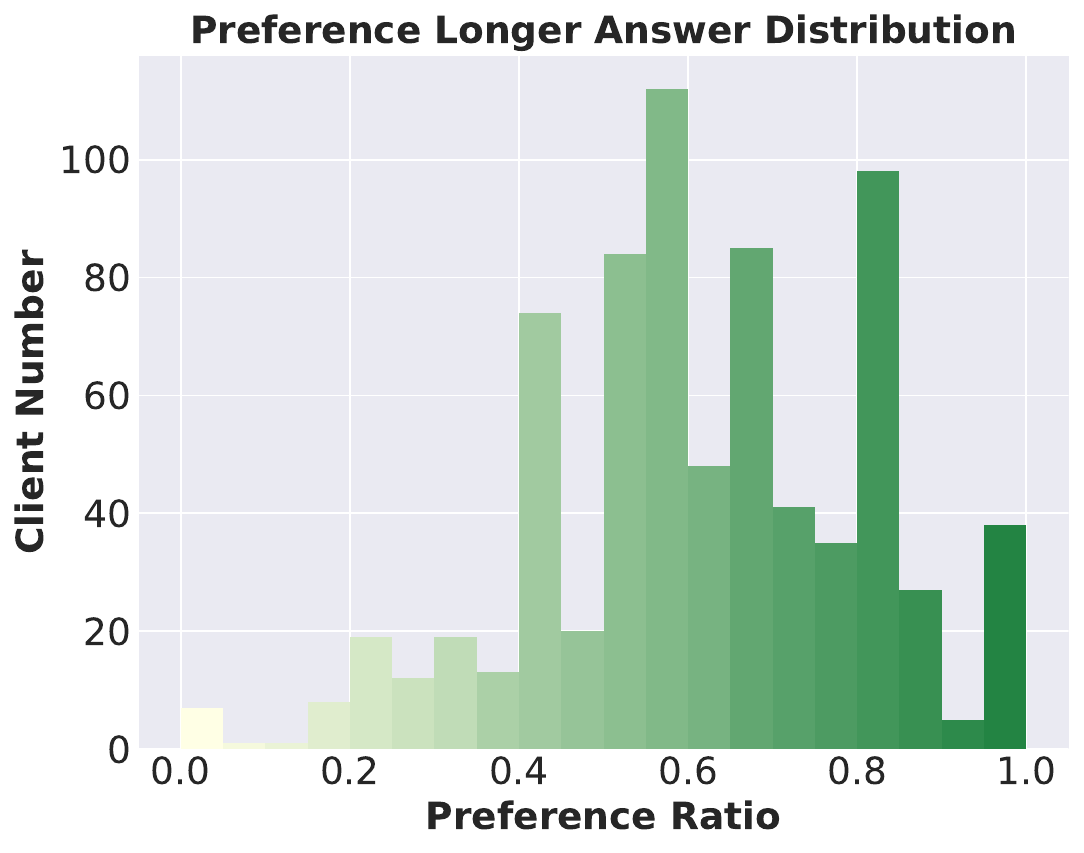}\label{fig:chatbotpa_length}
    }
    \subfigure[Quality Preference]{ 
    \centering
    \includegraphics[width=0.235\linewidth]{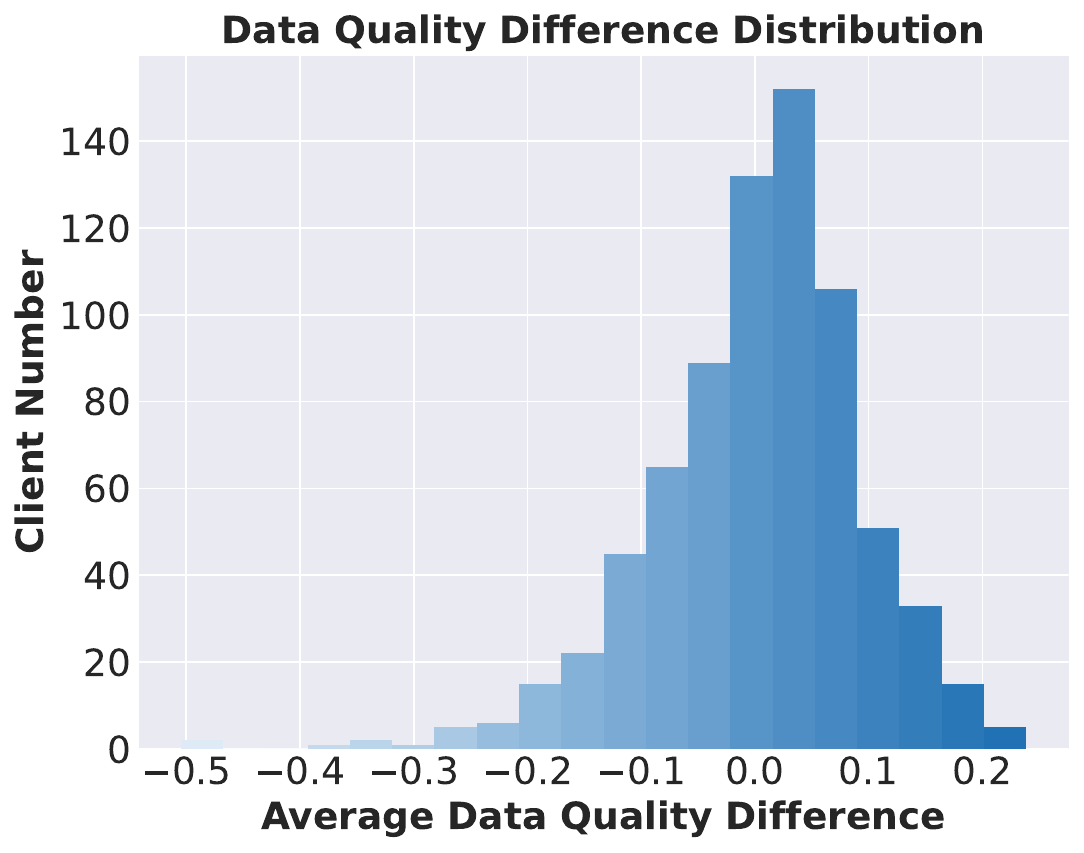}\label{fig:chatbotpa_quality}
    }
    \caption{(a) Langauge distribution of clients in Fed-Aya dataset. (b) The distribution of length of instruction and response of clients' data. (c) Distribution of length preference (the ratio of a user preferring longer response) of clients in Fed-ChatbotPA dataset. (d) Distribution of quality preference (quality difference between preferred and dispreferred data) of clients in Fed-ChatbotPA dataset.}
\end{figure}

\begin{figure}[t]
    \centering
    \subfigure[Fed-Aya]{
        \includegraphics[width=0.23\linewidth]{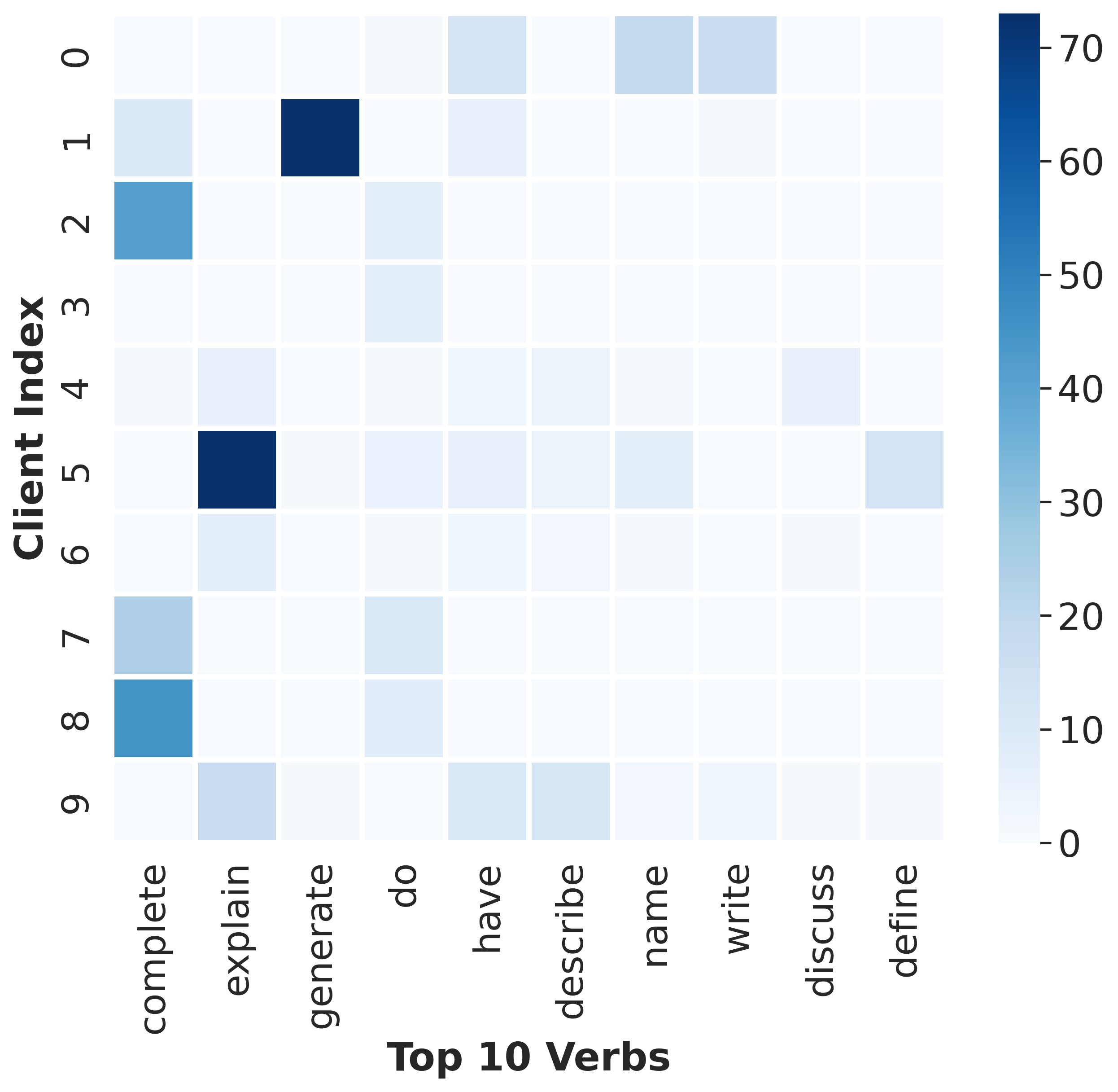}
            \label{fig:aya_heat}
        }
        \subfigure[Fed-ChatbotIT]{
        \includegraphics[width=0.23\linewidth]{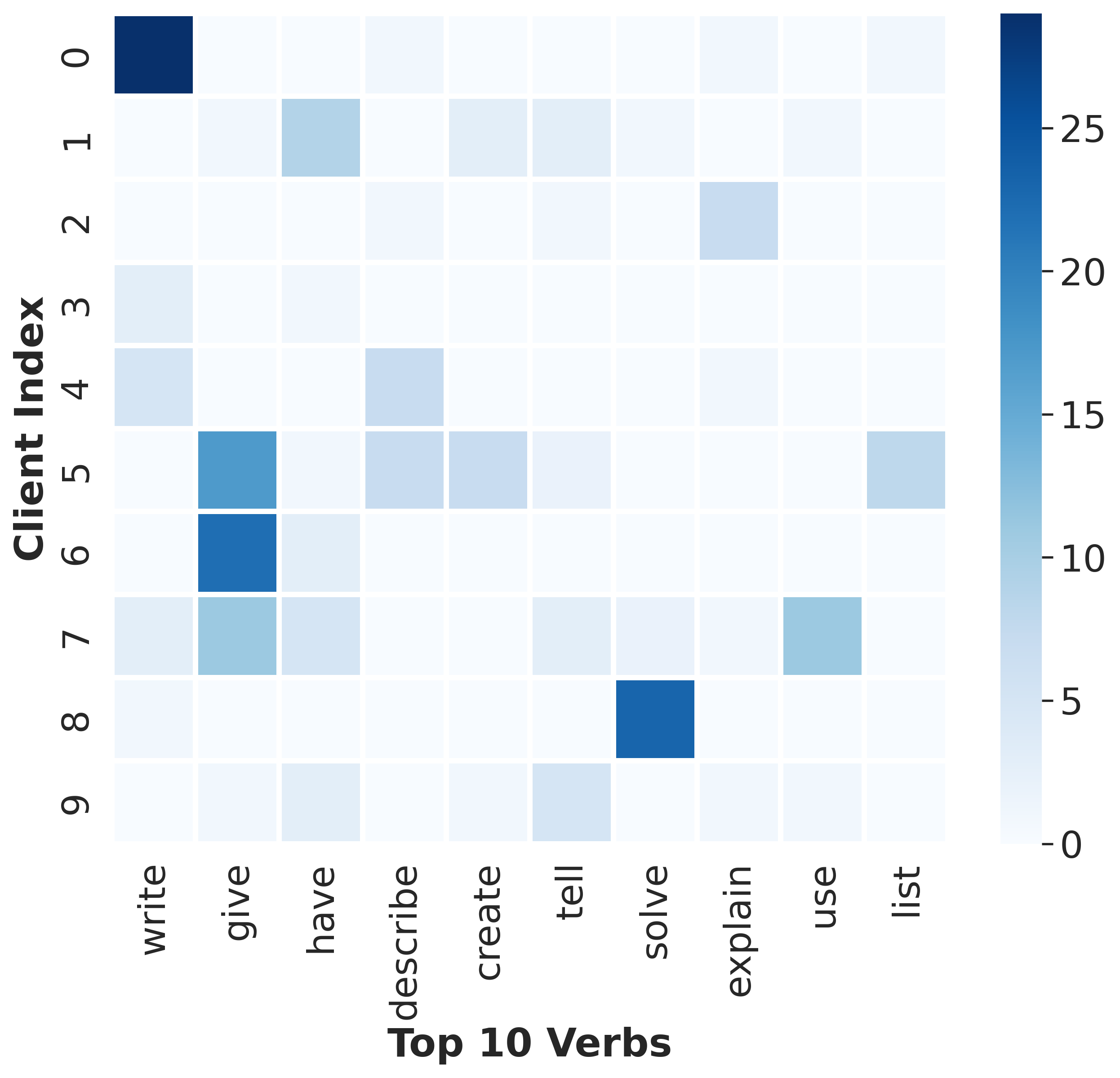}
            \label{fig:chatbot_heat}
        }
        \subfigure[Fed-WildChat]{
        \includegraphics[width=0.23\linewidth]{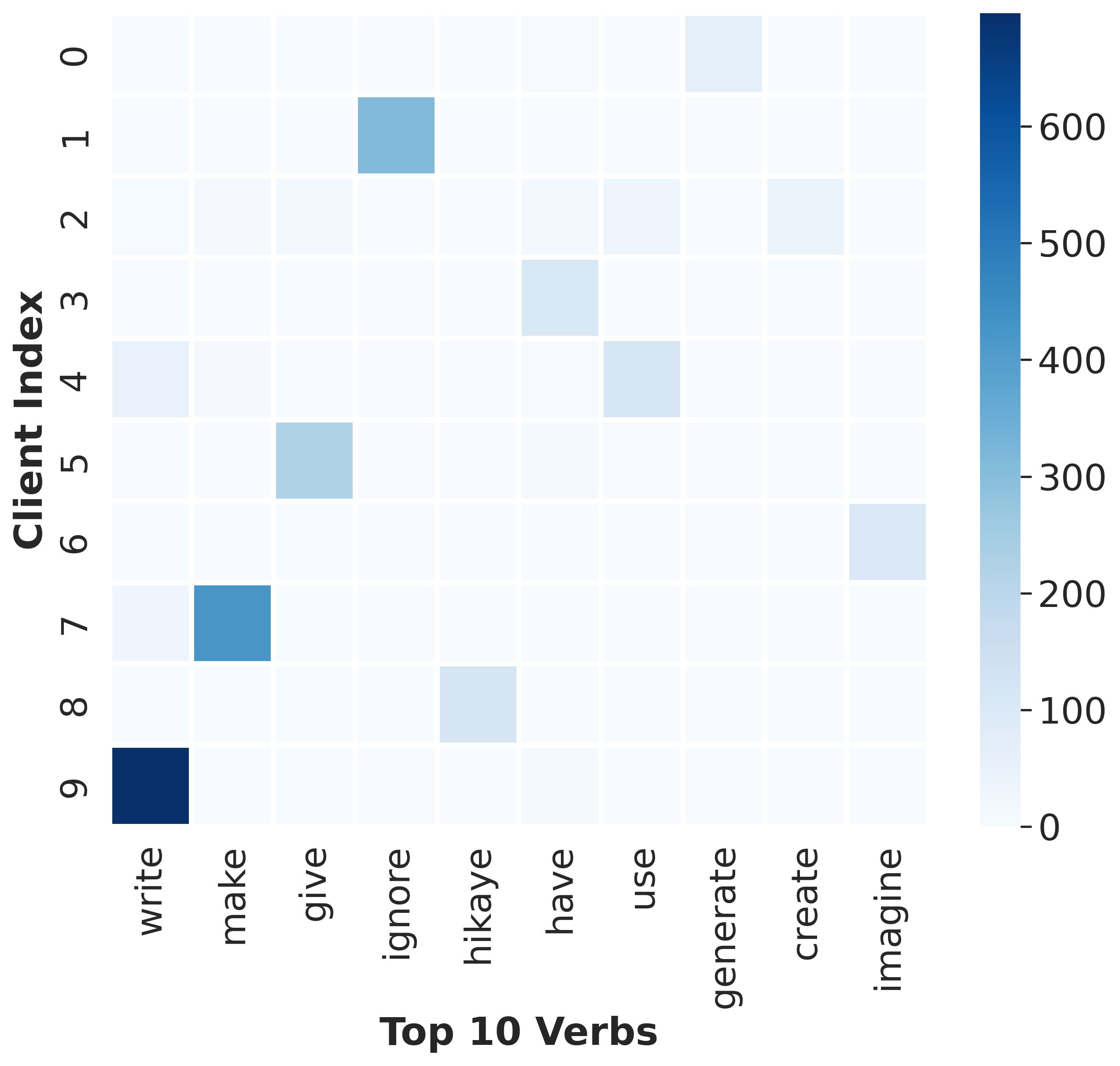}
            \label{fig:wildchat_heat}
        }
    \subfigure[Fed-ChatbotPA]{
    \includegraphics[width=0.23\linewidth]{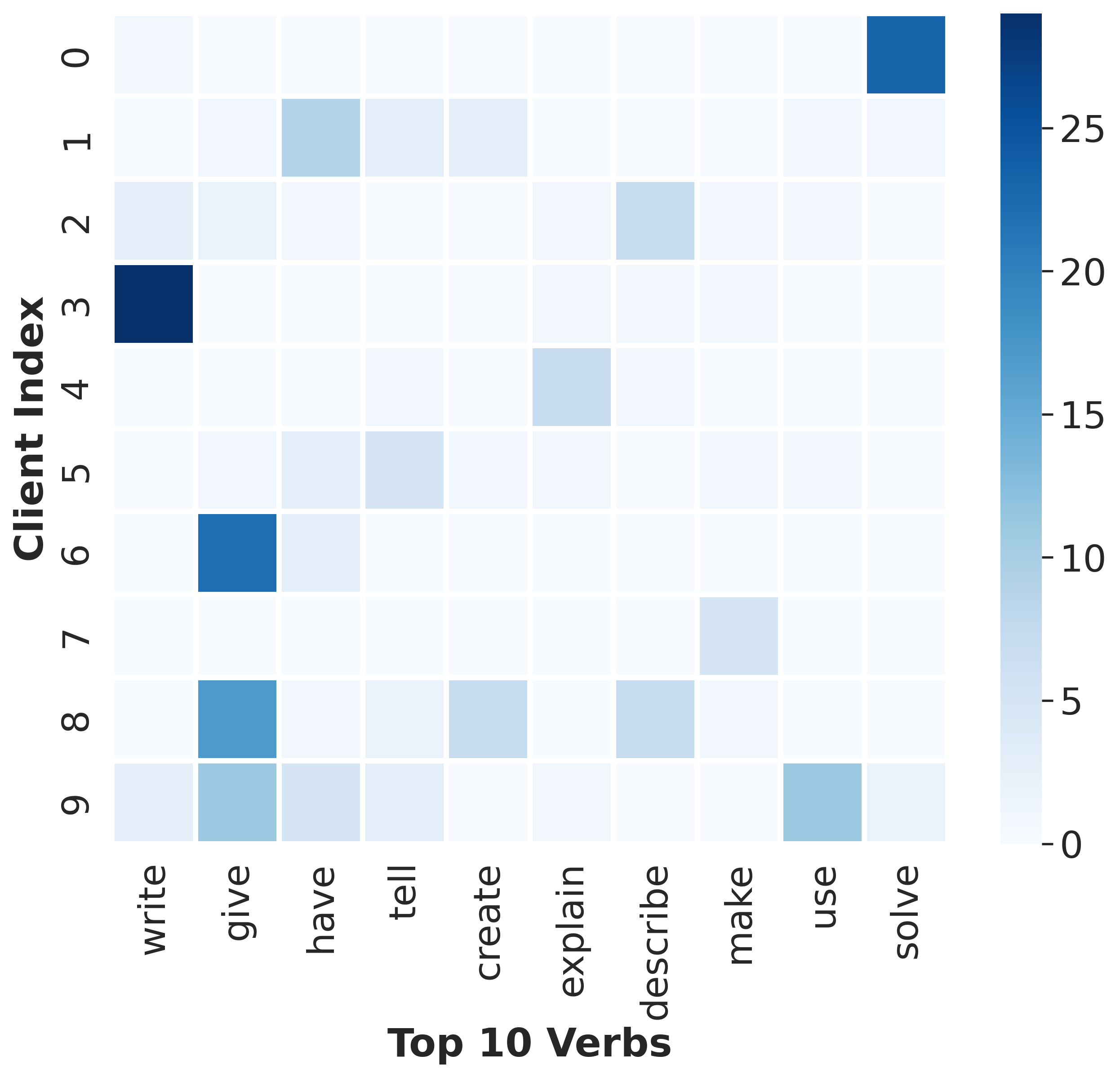}
            \label{fig:chatbot-dpo_heat}
        }
    \caption{Distributions of top 10 verbs in instructions (10 clients are plotted for illustration). Our realistic FedLLM datasets exhibit diverse patterns with respect to instruction types.}
    \label{fig:verb_heatmap}
\end{figure}

\begin{figure}[t]
\centering
\subfigure[Fed-Aya]{   
\begin{minipage}{3.2cm}
\centering    
\includegraphics[scale=0.18]{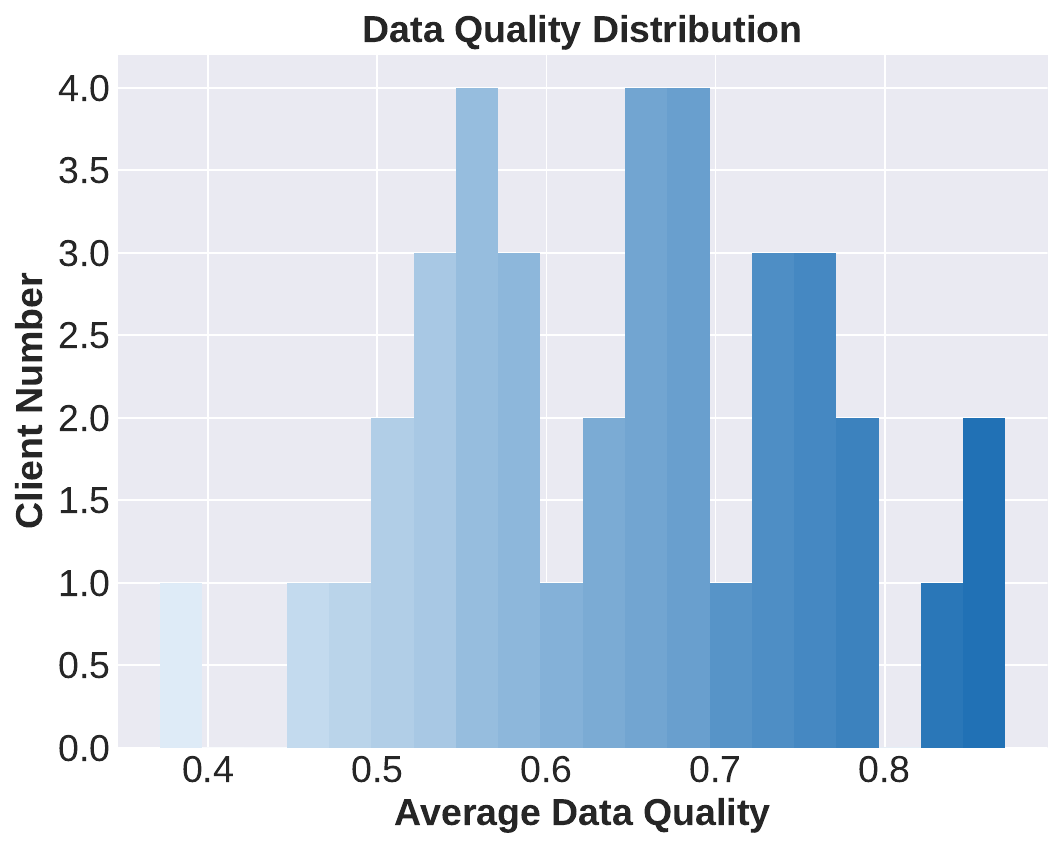}  
\end{minipage}
}
\subfigure[Fed-ChatbotIT]{ 
\begin{minipage}{3.2cm}
\centering
\includegraphics[scale=0.18]{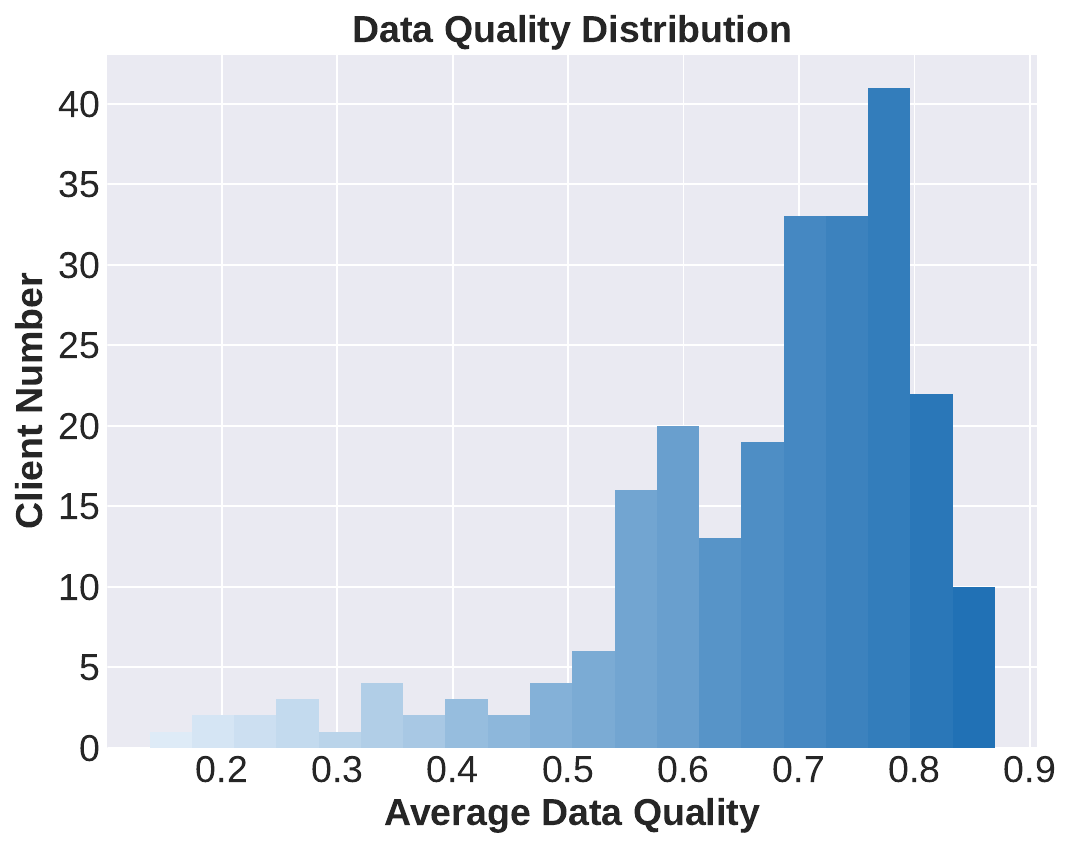}
\end{minipage}
}
\subfigure[Fed-WildChat]{ 
\begin{minipage}{3.2cm}
\centering
\includegraphics[scale=0.18]{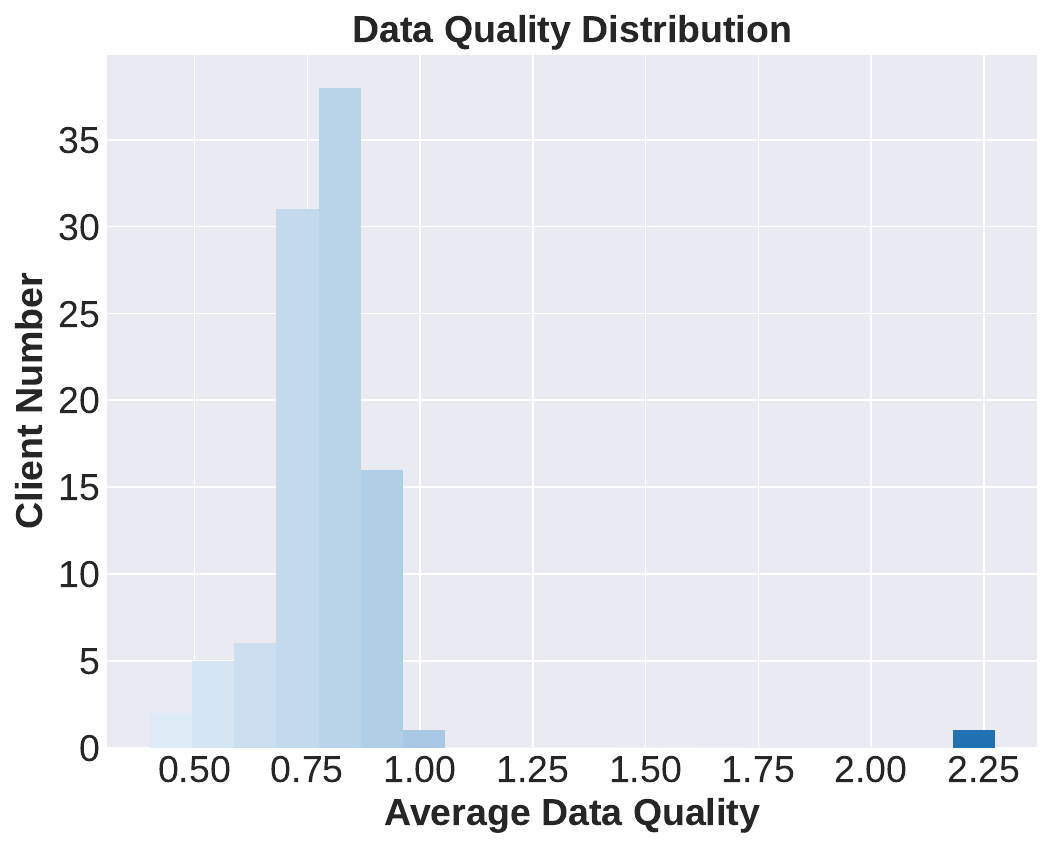}
\end{minipage}
}
\subfigure[Fed-ChatbotPA]{ 
\begin{minipage}{3.2cm}
\centering
\includegraphics[scale=0.18]{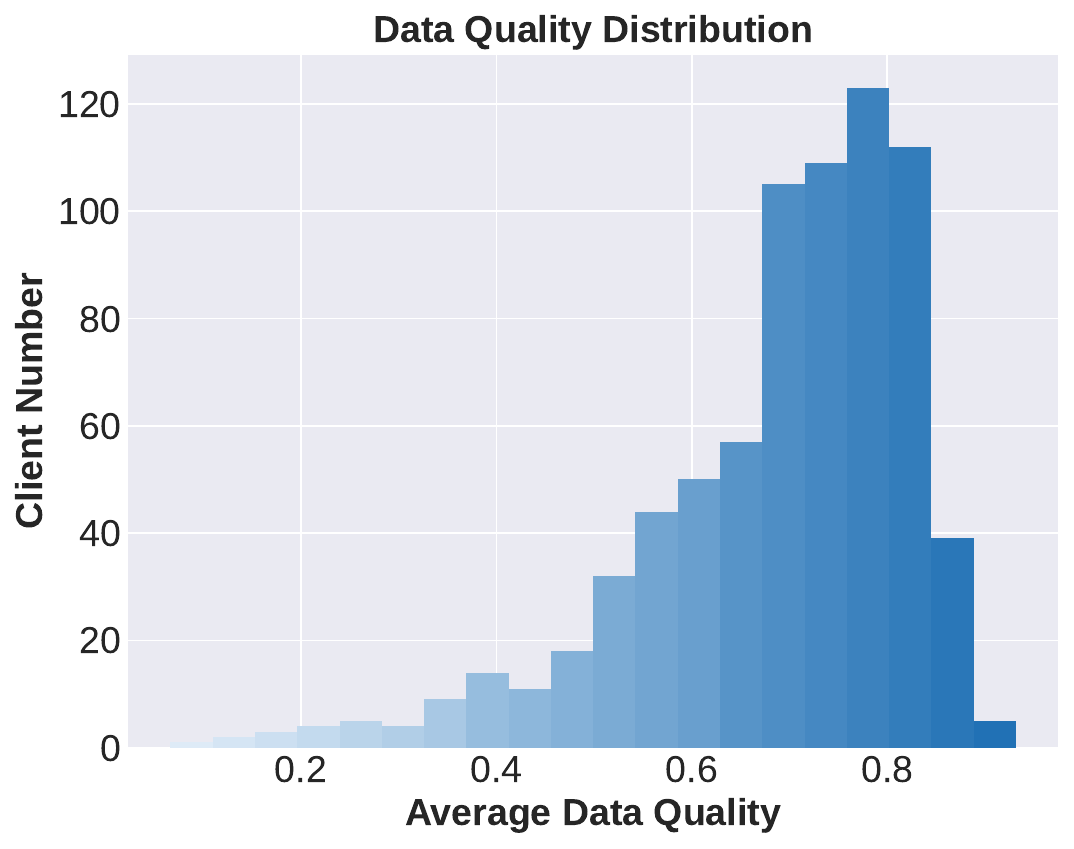}
\end{minipage}
}
\caption{The dataset quality distribution of clients in four training datasets: Fed-Aya, Fed-WildChat,
Fed-ChatbotIT and Fed-ChatbotPA.
We average the IFD scores of all instruction-response pairs of each client to represent the client's dataset quality.
}
\label{fig:quality_distribution}
\end{figure}

\begin{figure}[t]
\centering
\subfigure[Fed-Aya]{   
\begin{minipage}{3.2cm}
\centering    
\includegraphics[scale=0.16]{figs/tsne_aya.pdf}  
\end{minipage}
}
\subfigure[Fed-ChatbotIT]{ 
\begin{minipage}{3.2cm}
\centering
\includegraphics[scale=0.16]{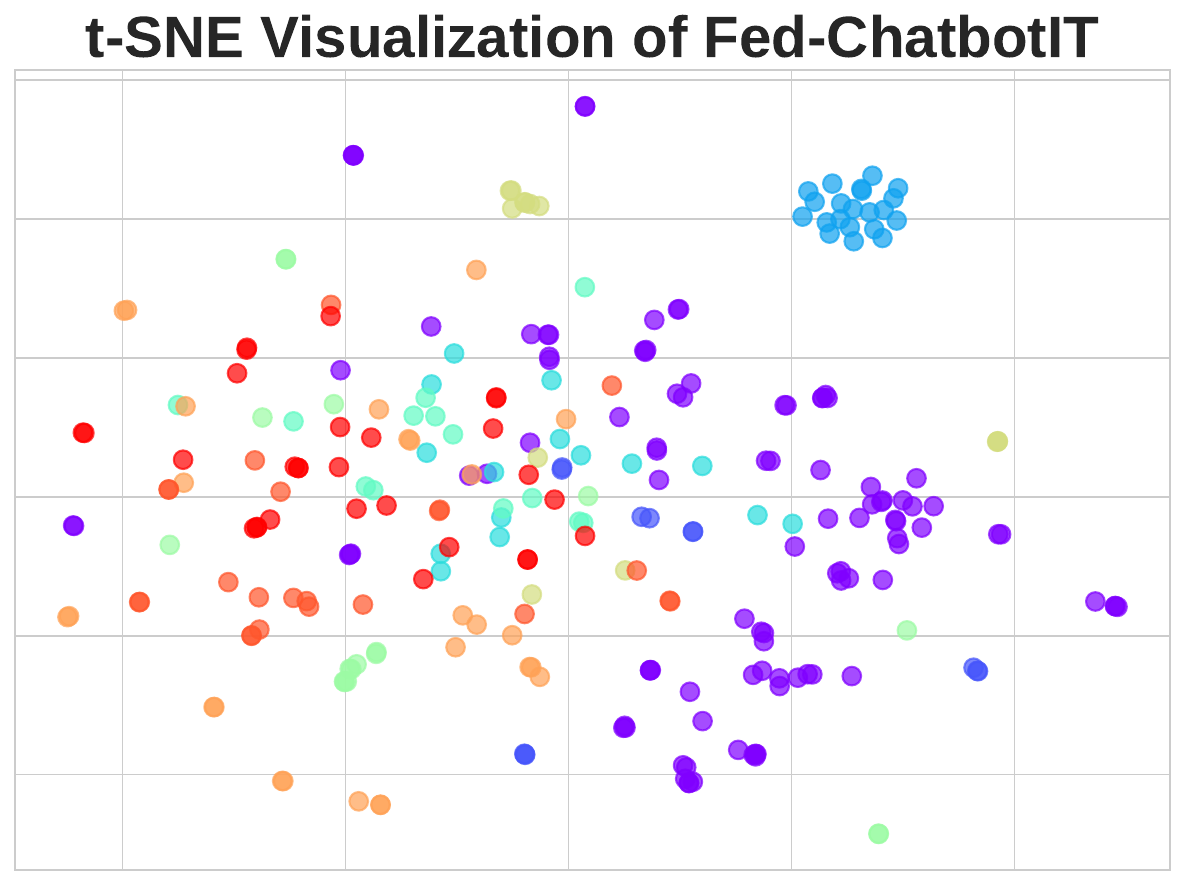}
\end{minipage}
}
\subfigure[Fed-WildChat]{ 
\begin{minipage}{3.2cm}
\centering
\includegraphics[scale=0.16]{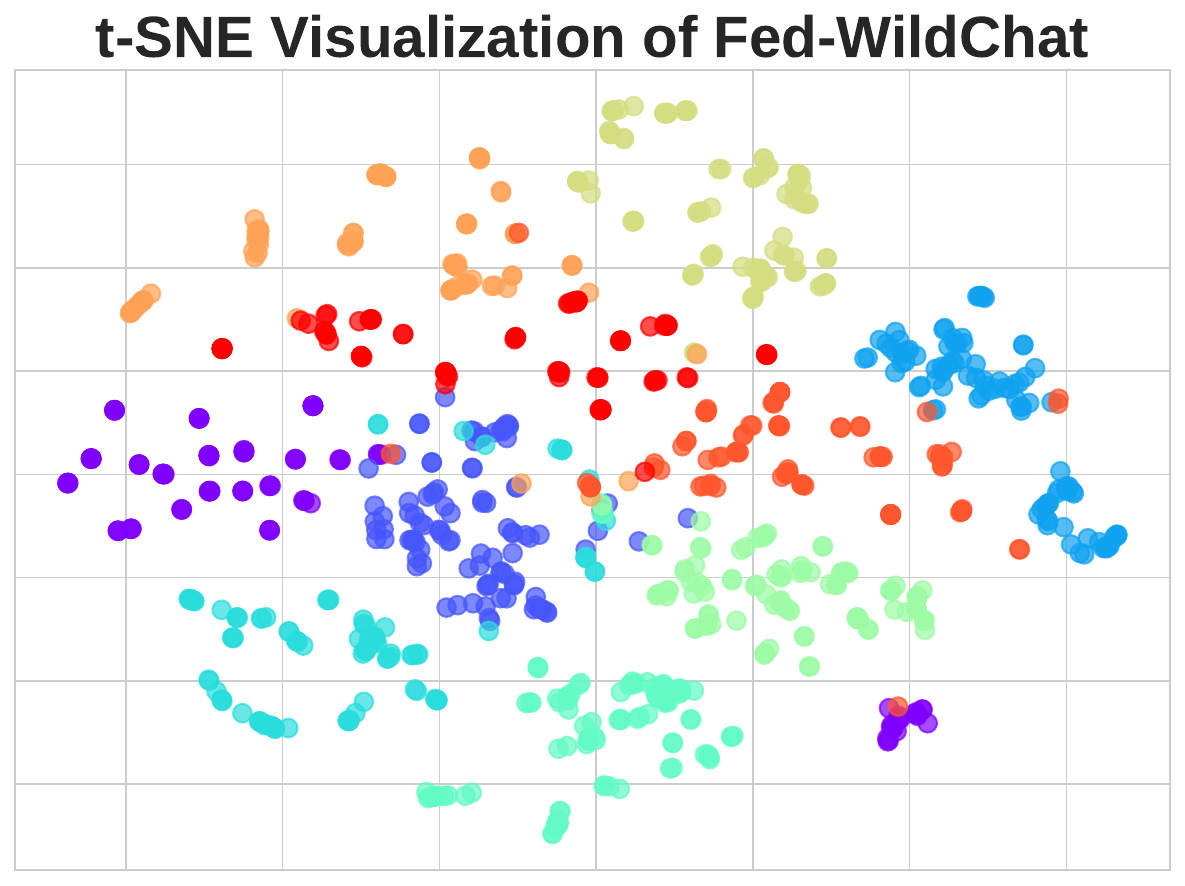}
\end{minipage}
}
\subfigure[Fed-ChatbotPA]{ 
\begin{minipage}{3.2cm}
\centering
\includegraphics[scale=0.16]{figs/tsne_chatbotPA.pdf}
\end{minipage}
}
\caption{The t-SNE visualization of embeddings of instruction-response pairs from 10 clients in Fed-Aya, Fed-ChatbotIT, Fed-WildChat, and Fed-ChatbotPA datasets. Each color denotes one client. We can see clustering phenomenon of one client's data and that clients' data are diverse.
}
\label{fig:tsne}

\end{figure}

\begin{figure}[t]
    \centering
    \subfigure[Fed-Aya]{\includegraphics[width=.237\linewidth]{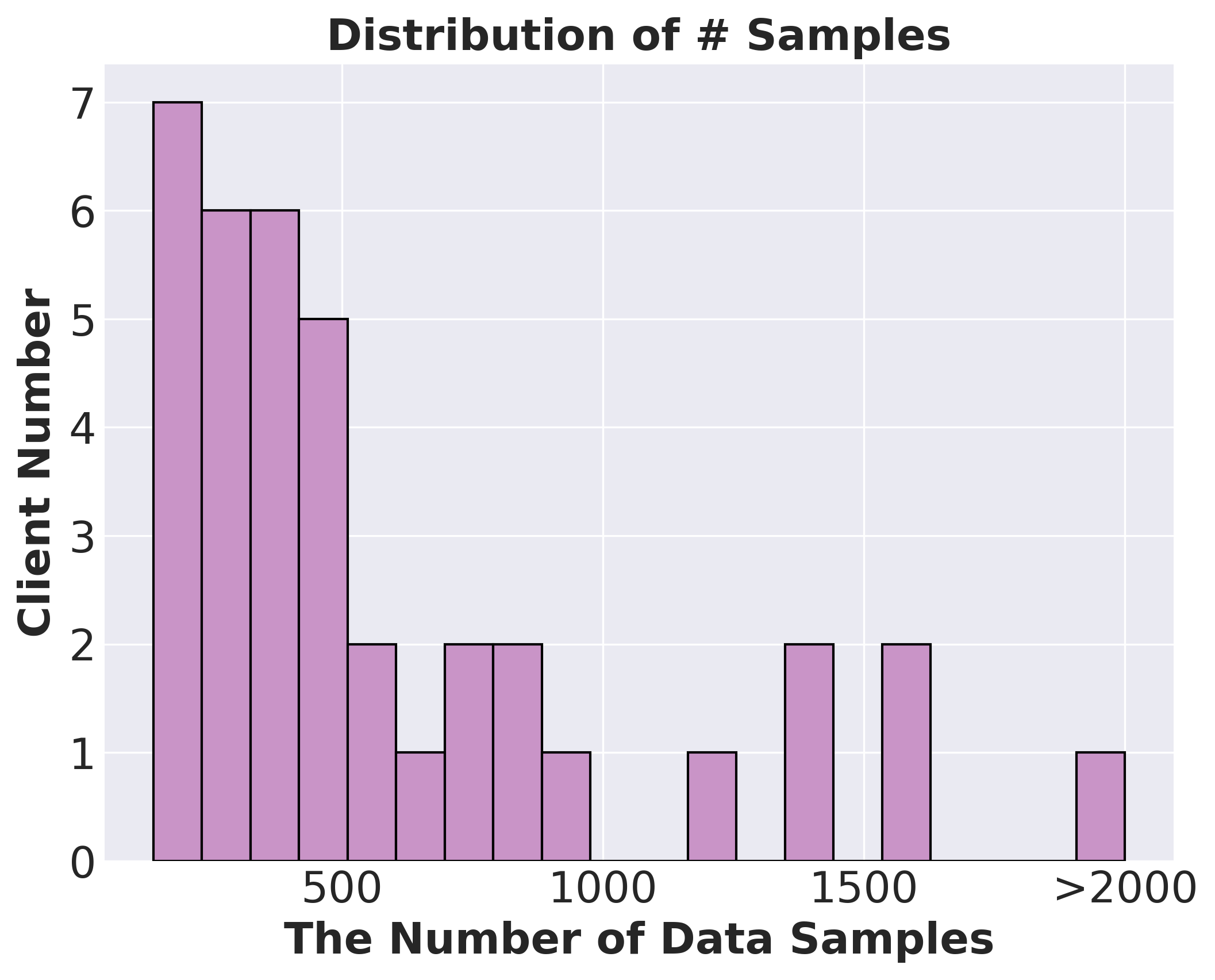}}
    \subfigure[Fed-ChatbotIT]{\includegraphics[width=.237\linewidth]{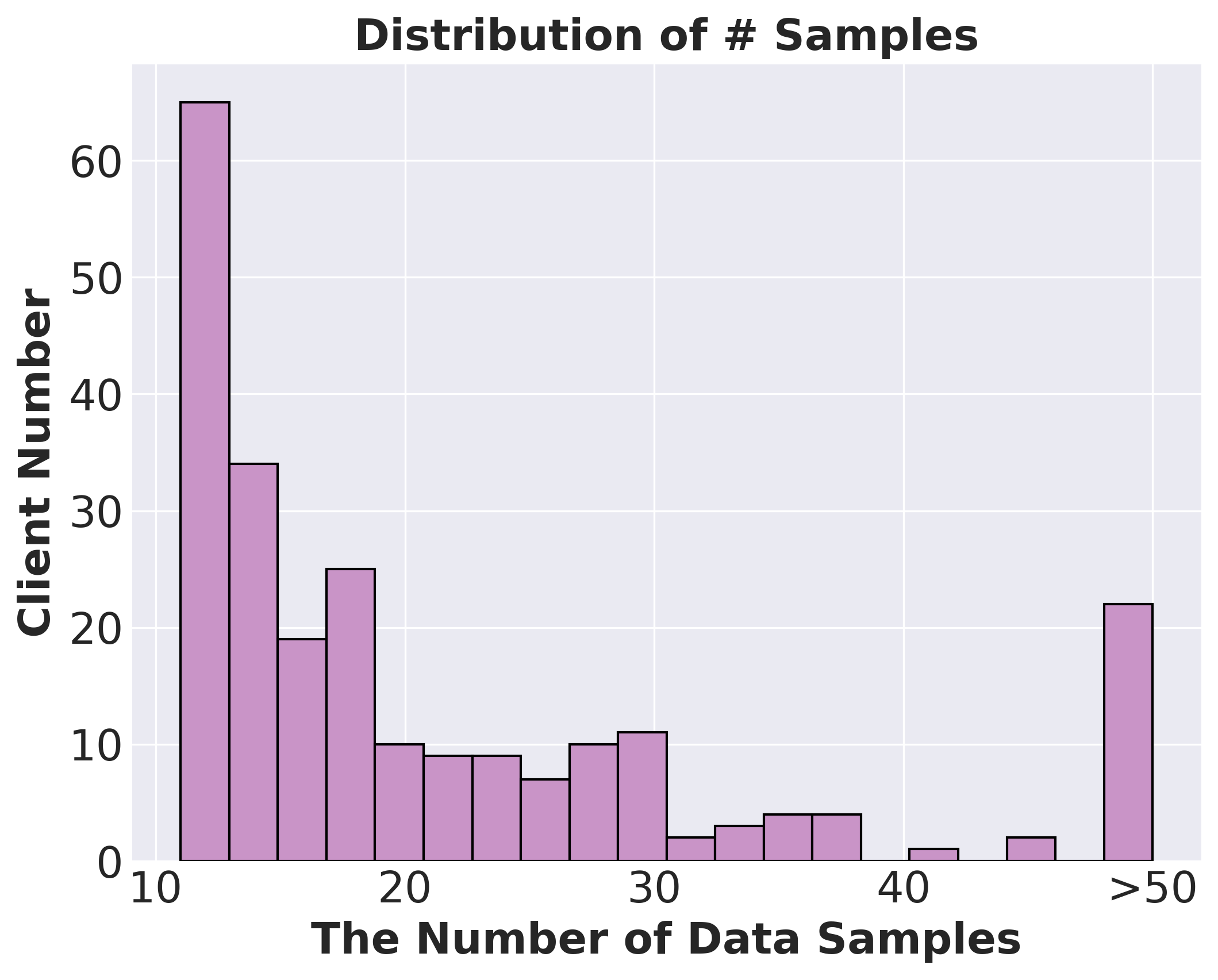}}
    \subfigure[Fed-WildChat]{\includegraphics[width=.237\linewidth]{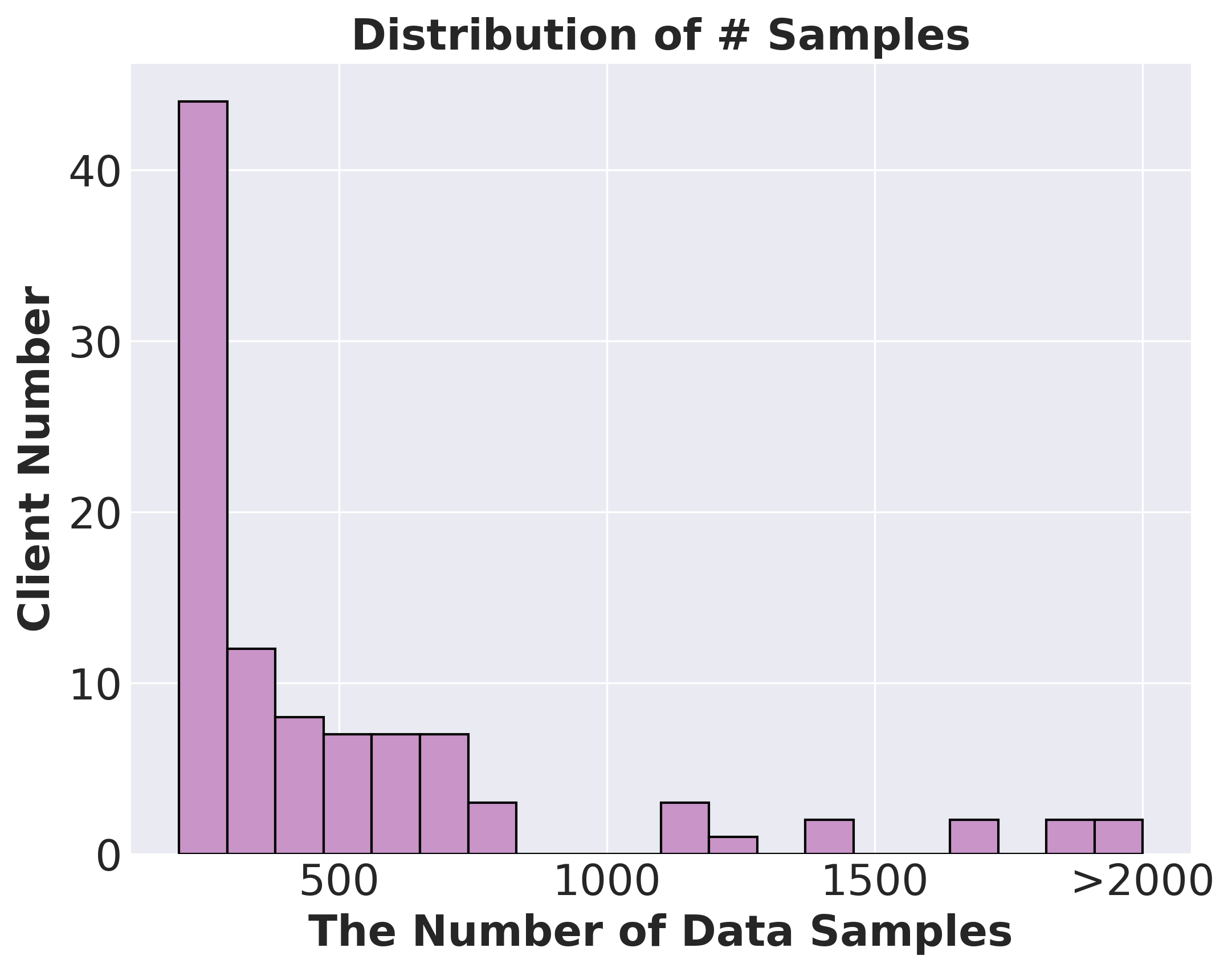}}
    \subfigure[Fed-ChatbotPA]{\includegraphics[width=.237\linewidth]{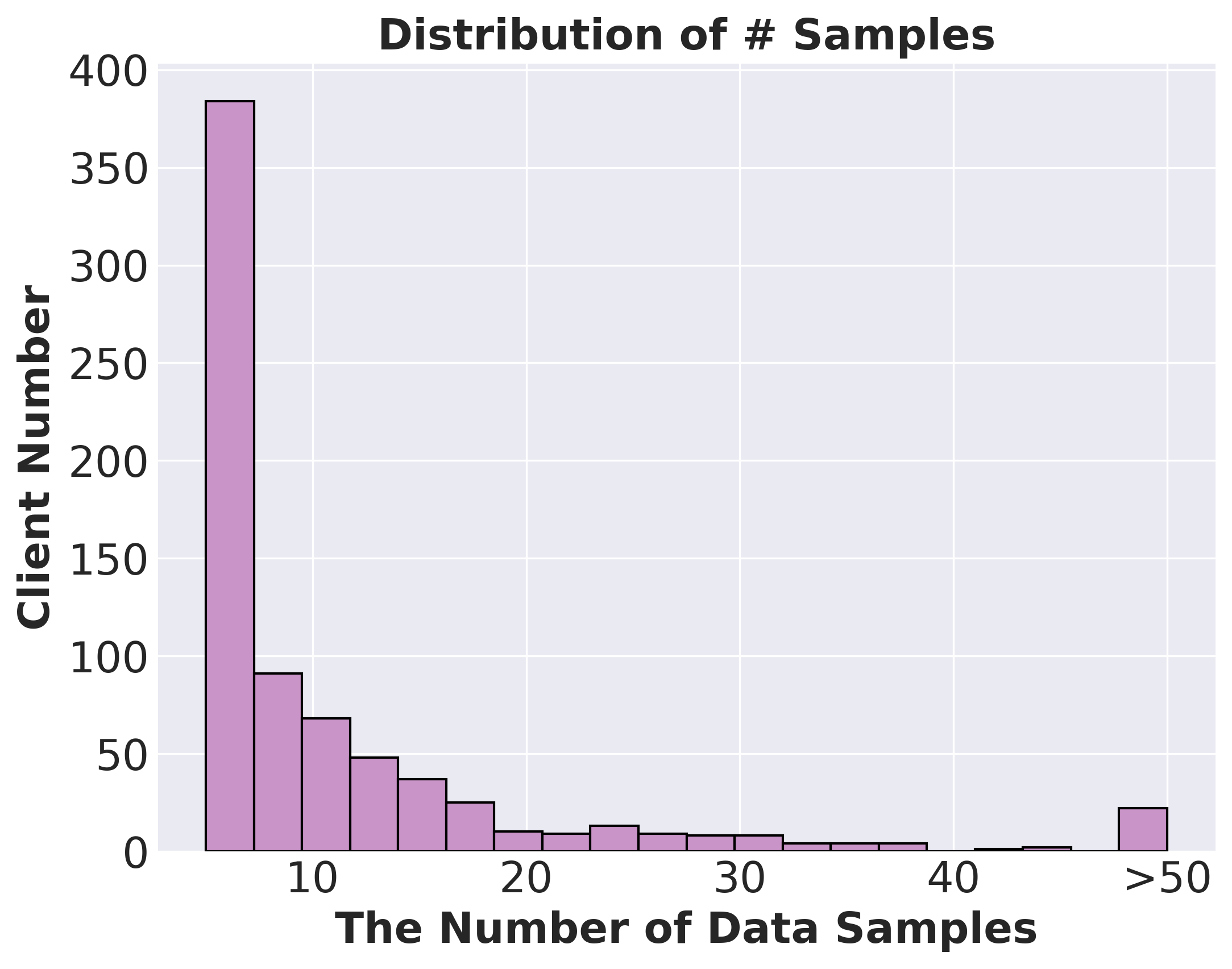}}
    \caption{Data quantity distribution across clients of our four FedLLM datasets. We can see a variety of data quantities of clients, where a large proportion of clients have relatively few data.}
    \label{fig:quantity_distribution}
\end{figure}

\textbf{Fed-Aya.}
Aya~\cite{singh2024aya} dataset is a multilingual instruction tuning dataset annotated by contributors from various countries~\cite{aya}.
We select 6 high-resource languages: English (en), Spanish (es), French (fr), Russian (ru), Portuguese (pt), Chinese (zh), and 2 low-resource languages: Standard Arabic (ar) and Telugu (te).
According to the annotator ID, we filter out those who contribute less than 100 annotations, and construct Fed-Aya, which consists of 38 clients with 25k data samples in total.
This dataset models a real-world federated scenario~\cite{dayan2021federated} where collaborating clients are distributed around the globe and aim to advance multilingual LLMs~\cite{openai2023gpt4,team2023internlm}.
We visualize the language distribution of Fed-Aya dataset in Figure~\ref{fig:aya_pie}, showing that the number of clients for different languages varies.
Therefore, it also provides a dataset basis for the explorations of new research topics in FedLLM, including language personalization and fairness across high- and low-resource languages.

\textbf{Fed-ChatbotIT.}
Chatbot-Arena-Conversations~\cite{zheng2023judging} is originally a collection of human-annotated preference data, where each data sample consists of a user instruction, a user-chosen response and a user-rejected response.
Here, for each data sample, we combine the instruction and user-chosen response as an instruction-response pair.
Subsequently, according to the user ID of the annotator, we filter out those who contribute less than 10 data samples and construct Fed-ChatbotIT, which consists of 237 clients with 6k data samples in total.
This dataset captures the diversities of realistic use cases in single-turn query of LLMs, where instructions of different users could hold different patterns.

\textbf{Fed-WildChat.}
WildChat~\cite{zhao2024wildchat} is a collection of conversations between humans and ChatGPT, which contains a broad spectrum of user-chatbot interactions.
According to the IP address, we partition the whole dataset into several user datasets and filter out those with less than 200 samples, forming our Fed-WildChat.
Fed-WildChat consists of 100 clients with 53k data samples in total.
This dataset represents real-world use cases between humans and chatbots, which involve multi-turn interactions.

\textbf{Fed-ChatbotPA.}
We construct another federated version of Chatbot-Arena-Conversations~\cite{chatbot} for preference alignment tasks: Fed-ChatbotPA.
Specifically, we filter out users who contribute fewer than 5 preference pairs and the resulting Fed-ChatbotPA consists of 747 clients with 10k data samples in total.
Each data sample contains a user instruction, a preferred and dispreferred response.
This dataset exhibits real-world property that different individuals could have different preferences.
To verify this, we analyze the dataset from two perspectives.
Firstly, we visualize the length preferences of clients in Figure~\ref{fig:chatbotpa_length}, where for each client we compute the ratio of the preferred responses being longer than the dispreferred responses.
We see that most clients tend to prefer longer responses (i.e., the ratio is larger than 0.5) and clients have various preference ratios.
Secondly, we visualize clients' quality preferences in Figure~\ref{fig:chatbotpa_quality}, where for each client we compute the averaged quality difference between the preferred and dispreferred data.
We can see the diversity of clients' quality preferences.

\subsection{Analysis of training datasets}
\label{sec:analysis_data}

We further show the diversities of our datasets for FedLLM from four perspectives.

\textbf{(1) Length.} For each client, we tokenize the instruction and response of each data sample using tokenizer of Llama2~\cite{llama2}, and average their length respectively for each client.
We plot the length distribution of clients in Figure~\ref{fig:length_violin}.
We can see that clients' data varies in data length and different datasets exhibit different distributions, verifying both inter-dataset and intra-dataset diversities.
\textbf{(2) Instruction.}
Following Self-Instruct~\cite{self-instruct}, we use the Berkeley Neural Parser~\cite{kitaev2018constituency} to parse the instructions and extract the root verbs.
We randomly sample 10 clients for each dataset and visualize the distribution of top-10 verbs in Figure~\ref{fig:verb_heatmap}.
We can see that clients have different usage preferences in their instructions and that the top 10 verbs vary among datasets.
Please refer to more detailed visualizations in Figure~\ref{fig:pie_chart}, \ref{fig:aya_big_pie}, \ref{fig:chatbot_big_pie}, \ref{fig:wildchat_big_pie}, \ref{fig:chatbotPA_big_pie}.
\textbf{(3) Quality.}
We measure the data quality using IFD metric~\cite{li2023quantity}, where a higher value denotes higher instruction-following difficulty, and average the quality of all data samples of one client.
From Figure~\ref{fig:quality_distribution}, we can observe the diversities in clients' data, and distinct distributions among these four datasets.
\textbf{(4) Embedding.}
We randomly sample 10 clients, extract the embedding of each instruction-response data sample using text-embedding-ada-002 model from OpenAI, and plot them via t-SNE~\cite{tsne} in Figure~\ref{fig:tsne}, where each color denotes one client.
We can see that there is clustering phenomenon, indicating certain patterns within one client's data that mirror real-world cases.
It also demonstrates the diversity of clients' data since dots with different colors are located at different regions.
\textbf{(5) Quantity.}
We plot clients' data quantity distribution in Figure~\ref{fig:quantity_distribution}.
From the figure, we see a variety of data quantity across clients for all datasets.

These analysis evidently reveals the diversities among clients' datasets, making our datasets appropriate candidates for the FedLLM community since they can mirror the complexities in real-world scenarios.

\subsection{Evaluation metrics}

To evaluate the effectiveness of the training methods on our realistic FL datasets, we consider 6 evaluation metrics, including 4 open-ended metrics and 2 close-ended metrics.

\textbf{Open-ended evaluation.}
MT-Bench~\cite{zheng2023judging} is one of the most acknowledged evaluation metrics in the era of LLMs, which evaluates both one-turn and two-turn conversation capability.
Similarly, Vicuna bench~\cite{vicuna2023} evaluates one-turn instruction-following capability.
AdvBench~\cite{zou2023universal} evaluates the safety rate given unsafe instructions.
Additionally, we consider an in-domain evaluation metric termed Ref-GPT4, where we randomly sample 50 unseen data as the test set and use GPT-4 to rate the generated response given the ground-truth response as reference~\cite{2024aya,zheng2023judging}; see prompt in Figure~\ref{fig:ref_gpt4_prompt}.

\textbf{Close-ended evaluation.}
We consider two common close-ended evaluations~\cite{chia-etal-2024-instructeval,llama2,jiang2023mistral,openai2023gpt4,ye2024openfedllm}: MMLU~\cite{hendrycks2021measuring} (measuring knowledge of 57 subjects) and HumanEval~\cite{chen2021evaluating} (measuring capability of generating programs from docstrings).
We evaluate these two metrics mainly to ensure that fine-tuning will not compromise these capabilities acquired during pre-training.

\section{Experiments on FedLLM-Bench}
\label{section:experiments}

\textbf{Experimental setups.}
For instruction tuning task, we use Llama2-7B~\cite{llama2} as the base model and set the learning rate as $2e^{-5}$ with a batch size of 16.
For preference alignment task, we use Alpaca-7B~\cite{alpaca} as the base model and set the learning rate $1e^{-4}$ with a batch size of 8, as a large proportion of clients have fewer than 10 data samples.
We adopt 8-bit quantization on the base model and set the rank of LoRA~\cite{hu2021lora} as $16$.
The number of communication rounds is set to either 100 or 200 and only a small proportion of clients are sampled for each round.
We set the number of steps of local model training as 10 for most scenarios.
Please refer to details in Appendix~\ref{app:exp_setup}.

\begin{table}[t]
  \caption{Experiments on multilingual dataset Fed-Aya evaluated via Ref-GPT4. FL methods generally perform better than local training on average. However, FL methods can not ensure better performance on every language, implying the necessity for exploring language personalization techniques.}
  \label{tab:fed-aya}
  \centering
  \begin{tabular}{l|cccccccc|>{\columncolor{gray!10}}c}
    \toprule
    Algorithm & ar & en & es & fr & pt & ru & te & zh & Average \\
    \midrule
    Local Training (ar) & 2.55 & 7.55 & 4.85 & 5.10 & 3.95 & 4.55 & 1.55 & 3.35 & 4.18 \\
    Local Training (en) & 2.55 & 7.20 & 5.35 & 4.60 & \textbf{5.35} & 4.75 & 1.60 & 3.55 & 4.37 \\
    Local Training (es) & 1.90 & 7.80 & 5.55 & 5.60 & 4.50 & 5.20 &  1.30 & 5.05 & 4.62 \\
    Local Training (fr) & 1.85 & 7.90 & 4.75 & 4.20 & 4.25 & 5.05 & 1.30 & 3.95 & 4.16 \\
    Local Training (pt) & 1.95 & 5.95 & 4.20 & 5.45 & 3.85 & 5.15 & 1.55 & 3.95 & 4.01 \\
    Local Training (ru) & 1.60 & 7.80 & 6.05 & 4.80 & 4.00 & 4.50 & 1.75 & 4.90 & 4.43 \\
    Local Training (te) & 2.10 & 3.70 & 3.75 & 3.50 & 3.05 & 4.10 & 1.25 & 3.60 & 3.13 \\
    Local Training (zh)  & 2.30 & 8.10 & 5.45 & 5.80 & 4.80 & 4.30 & 1.60 & 4.95 & 4.66 \\
    \midrule
    FedAvg~\cite{fedavg} & 2.50 & 8.00 & 5.50 & 5.35 & 4.95 & 5.65 & \textbf{2.00} & 5.25 & 4.90 \\
    FedProx~\cite{fedprox} & \textbf{3.20} & 7.10 & 5.90 & \textbf{5.65} & 4.85 & 5.20 & 1.60 & \textbf{5.80} & 4.92 \\
    SCAFFOLD~\cite{scaffold} & 2.65 & 7.75 & \textbf{6.30} & 5.35 & 5.00 & \textbf{6.35} & 1.45 & 4.90 & \textbf{4.97} \\
    FedAvgM~\cite{fedavgm} & 3.00 & 7.80 & 5.35 & 5.00 & 5.30 & 5.65 & 1.90 & 5.00 & 4.86 \\
    FedYogi~\cite{fedopt} & 2.00 & \textbf{8.45} & 6.15 & 4.55 & 3.85 & 6.30 & 1.65 & 4.93 & 4.73 \\
    FedAdagrad~\cite{fedopt} & 2.50 & 7.85 & 5.15 & 5.25 & 4.45 & 5.75 & 1.55 & 5.50 & 4.75 \\
    FedAdam~\cite{fedopt} & 2.40 & 8.50 & 5.25 & 4.70 & 4.35 & 5.40 & 1.90 & 5.20 & 4.71 \\
    \bottomrule
  \end{tabular}
\end{table}

\subsection{Benchmark results}

\textbf{Fed-Aya.}
Here, we conduct experiments on Fed-Aya with Ref-GPT4 as the evaluation metric for 8 languages.
We run local training for each language (randomly sample one client for simplicity), and 7 federated methods.
From Table~\ref{tab:fed-aya}, we see that:
(1) Most FL methods can outperform local training on average, indicating the effectiveness of collaboration.
(2) No FL method can achieve comprehensive superiority in all languages, implying the necessity of future exploration of language personalization~\cite{li2021ditto,pfedgraph}.
(3) FedAvg and FedProx are the two most effective algorithms here.

\begin{table}[t]
  \caption{Experiments on single-turn chat dataset Fed-ChatbotIT. FL methods perform consistently better under \colorbox{red!10}{open-ended} instruction-following evaluations and comparably under \colorbox{blue!10}{closed-ended} knowledge evaluations compared to local training. Overall, FedAdagrad is the most effective.}
  \label{tab:fed-chatbotit}
  \centering
  \begin{tabular}{>{\columncolor{gray!10}}l|>{\columncolor{red!10}}c>{\columncolor{red!10}}c>{\columncolor{red!10}}c>{\columncolor{red!10}}c|>{\columncolor{blue!10}}c>{\columncolor{blue!10}}c}
    \toprule
    Algorithm & MT-Bench-1 & Vicuna & Ref-GPT4 & Average & HumanEval & MMLU\\
    \midrule
    Local Training & 3.73 & 6.78 & 4.49 & 5.00 & 13.41 & 46.31 \\
    \midrule
    FedAvg~\cite{fedavg} & 4.30 & 6.93 & \textbf{5.29} & \textbf{5.51} & 14.02 & 46.10 \\
    FedProx~\cite{fedprox} & 4.25 & 7.21 & 5.00 & 5.49 & 14.63 & 46.12 \\
    SCAFFOLD~\cite{scaffold} & 3.86 & 7.35 & 4.82 & 5.34 & 15.24 & 46.02\\
    FedAvgM~\cite{fedavgm} & \textbf{4.34} & 7.17 & 4.76 & 5.42 & 14.63 & 46.13 \\
    FedYogi~\cite{fedopt} & 4.13 & 7.20 & 5.00 & 5.44 & \textbf{15.85} & 46.24 \\
    FedAdagrad~\cite{fedopt} & 3.94 & \textbf{7.50} & 4.99 & 5.48 & \textbf{15.85} & \textbf{46.48}\\
    FedAdam~\cite{fedopt} & 3.88 & 7.32 & 5.02 & 5.41 & 14.57 & 46.10 \\
    \bottomrule
  \end{tabular}
\end{table}

\textbf{Fed-ChatbotIT.}
Here, we conduct experiments on Fed-ChatbotIT evaluated under 5 metrics.
We randomly sample two clients to run local training and average their evaluation results.
From Table~\ref{tab:fed-chatbotit}, we see that
(1) on open-ended evaluation, all FL methods consistently outperform local training, indicating the effectiveness of FL in enhancing the capability of instruction following.
(2) On closed-ended evaluation, FL methods perform better or are comparable to local training, indicating that FL training will not compromise LLMs' general capability.
(3) Across all metrics, the best scores are achieved by FL algorithms; On average, FedAdagrad~\cite{fedopt} achieves the best performance.

\textbf{Fed-WildChat.}
Here, we show two series of experiments based on Fed-WildChat: instruction tuning based on single-turn and multi-turn conversations, in Table~\ref{tab:fed-wildchat}.
For both experiments, we see that FL methods consistently outperform local training, verifying the effectiveness of collaboration.
For single-turn, we see that no FL method can dominate in all evaluation metrics;
while for multi-turn, we see that FedAvg~\cite{fedavg} consistently outperforms the best across metrics.

This is an interesting observation since the other baseline methods are shown to be effective in tackling data heterogeneity in other tasks such as image classification~\cite{fedprox,scaffold}.
This phenomenon could be attributed to two reasons:
(1) training from pre-trained model itself benefits tackling the issue of data heterogeneity~\cite{nguyen2023begin,chen2023importance}, which could make some model-level optimization techniques not as effective as before~\cite{fedprox,scaffold}.
(2) We are fine-tuning with parameter-efficient fine-tuning technique~\cite{hu2021lora} with a small number of local steps (e.g., 10), reducing the risk of overfitting on local datasets~\cite{scaffold,wangsurvey}.
Therefore, we call for more future works to enhance the performance regarding data, such as considering data quality~\cite{zhao2024enhancing} or synthetic data~\cite{zhang2024fedpit}.

\begin{table}[t]
  \caption{Experiments of \colorbox{red!10}{single-turn} and \colorbox{blue!10}{multi-turn} chat on Fed-Wildchat. FL methods perform consistently better than local training. FedAvg is a robust method in this scenario.}
  \label{tab:fed-wildchat}
  \centering
  \begin{tabular}{>{\columncolor{gray!10}}l|>{\columncolor{red!10}}c>{\columncolor{red!10}}c>{\columncolor{red!10}}c|>{\columncolor{blue!10}}c>{\columncolor{blue!10}}c>{\columncolor{blue!10}}c>{\columncolor{blue!10}}c}
    \toprule
    Experiment & \multicolumn{3}{>{\columncolor{red!10}}c|}{Single-Turn} & \multicolumn{4}{>{\columncolor{blue!10}}c}{Multi-Turn}\\
    Algorithm & MT-1 & Vicuna & Ref-GPT4 & MT-1 & MT-2 & MT-Bench & Ref-GPT4 \\
    \midrule
    Local Training & 4.15 & 7.03 & 4.50 & 3.99 & 2.56 & 3.27 & 4.68 \\
    \midrule
    FedAvg~\cite{fedavg} & 4.81 & 7.99 & 5.88 & \textbf{4.84} & \textbf{3.15} & \textbf{3.99} & \textbf{5.86} \\
    FedProx~\cite{fedprox} & \textbf{4.86} & 7.93 & 5.74 & 4.58 & 2.92 & 3.75 & 5.26 \\
    SCAFFOLD~\cite{scaffold} & 4.78 & 7.93 & 5.57 & 4.46 & 3.13 & 3.79 & 5.25 \\
    FedAvgM~\cite{fedavgm} & 4.52 & \textbf{8.07} & 5.85 & 4.53 & 2.77 & 3.65 & 5.34 \\
    FedYogi~\cite{fedopt} & 4.78 & 8.04 & 5.48 & 4.59 & 2.96 & 3.78 & 5.05 \\
    FedAdagrad~\cite{fedopt} & 4.76 & 7.76 & \textbf{5.93} & 4.64 & 3.03 & 3.84 & 5.17 \\
    FedAdam~\cite{fedopt} & 4.54 & 8.03 & 5.68 & 4.63 & 2.85 & 3.74 & 4.96 \\
    \bottomrule
  \end{tabular}
\end{table}

\textbf{Fed-ChatbotPA.}
Here, we conduct experiments of federated preference alignment on Fed-ChatbotPA dataset, with an instruction-tuned LLM as the model initialization.
We randomly sample two clients to run local training and average their evaluation results.
From Table~\ref{tab:fed-chatbotpa}, we see that
(1) preference alignment could enhance the LLMs' capability in following humans instructions in an helpful and harmless manner.
(2) All FL methods consistently perform better than local training, indicating the effectiveness of federated preference alignment.
Since the high-quality preference data usually involves massive human efforts, each party is hard to scale up the data, motivating diverse parties to collaborate via FL~\cite{ji2024beavertails,bai2022training,ganguli2022red}.
(3) Regarding instruction-following capabilities, FedAvgM~\cite{fedavgm}, FedProx~\cite{fedprox}, SCAFFOLD~\cite{scaffold}, adn FedAvg~\cite{fedavg} are four most effective methods.

\begin{table}[t]
  \caption{Experiments of federated preference alignment on Fed-ChatbotPA dataset. FL methods consistently perform better than local training, indicating the significance of collaboration via FL. Compared to base model, models trained via FL methods achieve consistent improvement in \colorbox{orange!10}{instruction-following} capabilities and \colorbox{red!10}{safety}, and preserve most of the \colorbox{blue!10}{knowledge}.}
  \label{tab:fed-chatbotpa}
  \centering
  \begin{tabular}{>{\columncolor{gray!10}}l|>{\columncolor{orange!10}}c>
  {\columncolor{orange!10}}c>{\columncolor{orange!10}}c|>{\columncolor{red!10}}c|>{\columncolor{blue!10}}c}
    \toprule
    Algorithm & MT-Bench-1 & Vicuna & Average & AdvBench & MMLU \\
    \midrule
    Base Model & 3.96 & 6.31 & 5.14 & 9.40 & \textbf{40.41} \\
    \midrule
    Local Training & 4.12 & 6.62 & 5.37 & 11.0 & 38.26 \\
    \midrule
    FedAvg~\cite{fedavg} & 4.44 & 7.06 & 5.75 & \textbf{16.2} & 39.70 \\
    FedProx~\cite{fedprox} & 4.44 & \textbf{7.11} & 5.78 & 13.8 & 39.51 \\
    SCAFFOLD~\cite{scaffold} & 4.53 & 7.01 & 5.77 & 16.0 & 39.94 \\
    FedAvgM~\cite{fedavgm} & \textbf{4.71} & 6.87 & \textbf{5.79} & 13.3 & 39.78 \\
    FedYogi~\cite{fedopt} & 4.33 & 6.62 & 5.48 & 11.3 & 40.27 \\
    FedAdagrad~\cite{fedopt} & 4.40 & 6.79 & 5.60 & 11.0 & 40.30 \\
    FedAdam~\cite{fedopt} & 4.31 & 6.72 & 5.52 & 11.8 & 40.26 \\
    \bottomrule
  \end{tabular}
\end{table}

\subsection{Further explorations}

\begin{table}[t]
  \caption{Experiments of exploration of efficient collaboration among languages. FedSimLang performs better than FedSamLang on some languages, indicating its partial effectiveness and calling for future works on constructing efficient collaboration structure to facilitate multilingual collaboration.}
  \label{tab:explore_language}
  \centering
  \begin{tabular}{l|ccccccc|>{\columncolor{gray!10}}c}
    \toprule
    Algorithm & ar & es & en & fr & pt & ru & zh & Average \\
    \midrule
    Local & 2.55 & 5.55 & 7.20 & 4.20 & 3.85 & 4.50 & 4.95 & 4.69 \\
    FedAvg  & 2.50 & 5.50 & \textbf{8.00} & 5.35 & \textbf{4.95} & \textbf{5.65} & 5.25 & 5.31 \\
    FedSamLang & \textbf{3.30} & \textbf{5.90} & 7.65 & \textbf{6.45} & 4.10 & 4.80 & 5.35 & \textbf{5.36} \\
    FedSimLang & 3.05 & 5.85 & 7.80 & 5.40 & 4.90 & 4.30 & \textbf{5.75} & 5.30 \\
    \bottomrule
  \end{tabular}
\end{table}

\textbf{Multilingual collaboration.}
We have observed in Table~\ref{tab:fed-aya} that despite that FL methods achieve better performance than local training on average, they fail to bring consistent benefits on every specific language.
Such observation motivates us to explore language personalization.
Therefore, in this experiment, we construct two representative baselines: FedAvg among clients with the same language (FedSamLang) and FedAvg among clients with "similar" languages (FedSimLang) to explore the potential mutual benefits among languages.
We partition languages into five "similar" groups by their language family~\cite{atkinson2006old} as follows:
(1) Standard Arabic, Urdu, and Iranian Persian;
(2) French, Italian, Spanish, and Portuguese;
(3) English and German;
(4) Russian, Polish and Ukrainian;
(5) Simplified Chinese, Traditional Chinese, Japanese and Korean.

We show the experimental results in Table~\ref{tab:explore_language}, where we compare FedSamLang and FedSimLang with FedAvg (trained on 8 languages as previous experiments) and local training.
From the results, we can see that
(1) FedSamLang outperforms local training in all languages and achieves the highest average score, indicating the benefits of collaboration among clients with the same language.
(2) Compared to FedSamLang, FedSimLang performs better on 3 languages (i.e., en, pt, and zh) but worse on other languages, showing that leveraging the power of other languages can benefit some particular languages.
Though this observation verifies the possibility of multilingual collaboration, we need more future works to fully explore its potential.
(3) FedSamLang and FedSimLang perform better or comparably compared to FedAvg with fewer collaborators, indicating the effectiveness of language personalization.
These results call for future works on exploring personalization techniques that can strike a good balance between localization and collaboration or construct a better collaboration structure among these multilingual clients~\cite{pfedgraph}.

\begin{wrapfigure}{r}{5.6cm}
   \centering
   % \vspace{-3mm}
    \includegraphics[width=0.4\columnwidth]{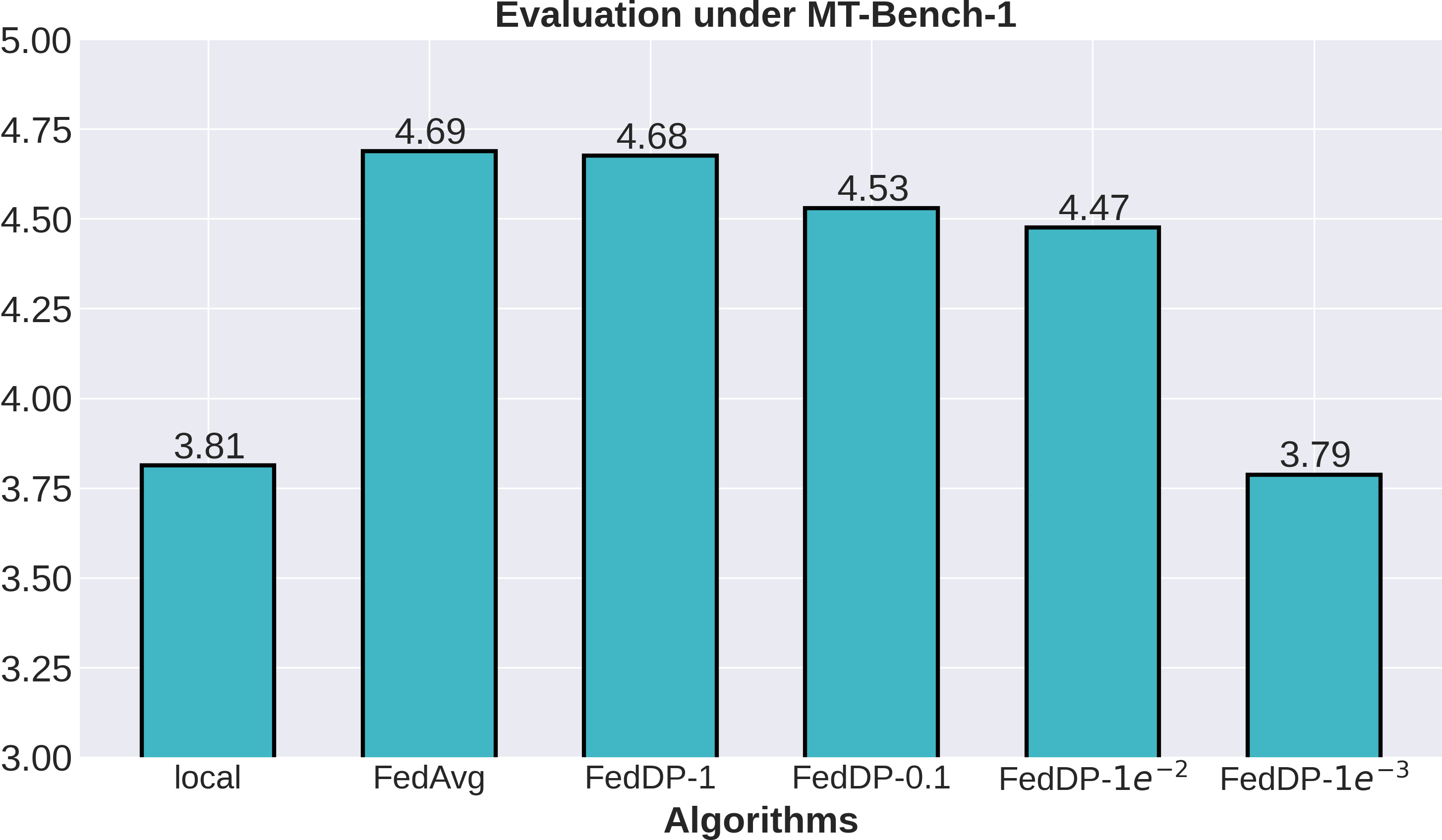}
    \vspace{-3mm}
    \caption{Experiments on Fed-WildChat with differential privacy $(\epsilon,\delta)$ ($\delta=1e^{-4}$). FedDP-x indicates $\epsilon=x$. FedAvg with $(1e^{-2},1e^{-4})$-DP still outperforms local training without DP while ensuring user-level differential privacy.}
    \vspace{-3mm}
    \label{fig:wildchat_dp}
\end{wrapfigure}
\textbf{Differential privacy.}
Here, we conduct experiments to evaluate the effectiveness of differential privacy~\cite{dwork2006calibrating}, where we apply user-level differential privacy~\cite{DBLP:journals/corr/abs-2003-00229}.
Experiments are conducted on our Fed-WildChat single-turn dataset; see more details in Appendix~\ref{app:dp}.
We fix the $\delta=1e^{-4}$ and tune $\epsilon$ in range of $\{1e^{-3}, 1e^{-2}, 0.1, 1\}$ that satisfies $(\epsilon,\delta)$-differential privacy, and report the results in Figure~\ref{fig:wildchat_dp}.
Results show that
(1) FedAvg with $(1,1e^{-4})$-differential privacy can achieve comparable performance compared to FedAvg without differential privacy.
(2) With the reduction of $\epsilon$, the privacy preservation improves while the performance degrades.
FedAvg with $(1e^{-3},1e^{-4})$-differential privacy can achieve comparable performance compared to local training without differential privacy technique.
To the best of our knowledge, this is the first time in the literature demonstrating the results of differential privacy in FedLLM.

\section{Conclusions}

Federated learning enables multiple parties to collaboratively train large language models without sharing their raw data (FedLLM), which has attracted many research efforts from the community.
In this paper, we propose the first realistic benchmark for FedLLM, FedLLM-Bench, which involves 8 training methods, 4 training datasets, and 6 evaluation metrics.
The core contribution lies in the datasets, which cover a wide range of client scale and two common tasks (i.e., instruction tuning and preference alignment).
These datasets exhibit many real-world diversities, including language, quality, quantity, instruction, sequence length, embedding, and preference, mirroring real-world scenarios.
Based on FedLLM-Bench, we conduct extensive experiments on all datasets to benchmark classical FL methods.
Besides, we also conduct experiments to explore the effective collaboration strategies of cross-language collaboration and show results when incorporating differential privacy with federated instruction tuning.
We believe that our FedLLM-Bench could benefit the FedLLM community by reducing required efforts, providing a practical testbed, and promoting fair comparisons.

\newpage
{
\small
\bibliographystyle{unsrt}
\bibliography{ref}
}

\medskip

%%%%%%%%%%%%%%%%%%%%%%%%%%%%%%%%%%%%%%%%%%%%%%%%%%%%%%%%%%%%

\newpage
\appendix

\section{Limitations}
\label{app:limitations}

Firstly, we explore Llama-2~\cite{llama2} for instruction tuning task and Alpaca~\cite{alpaca} for preference alignment task.
More future works are required to explore more model series and sizes.
Secondly, safety alignment is also an important topic in the era of LLMs, which is not comprehensively covered in our paper.
This could be an interesting and promising future direction.

\section{Datasets}

\subsection{Lengths measurement}
To measure each data sample, We use Llama2 tokenizer to tokenize the instruction and response of each data sample and use the number of tokens as the sentence length. Figure~\ref{fig:length_violin} shows the distribution of length of instruction and response of clients' data of our four datasets.

\subsection{Verbs and nouns}
\begin{figure}[h]
    \centering
    \subfigure[Aya]{
    \includegraphics[width=.48\linewidth]{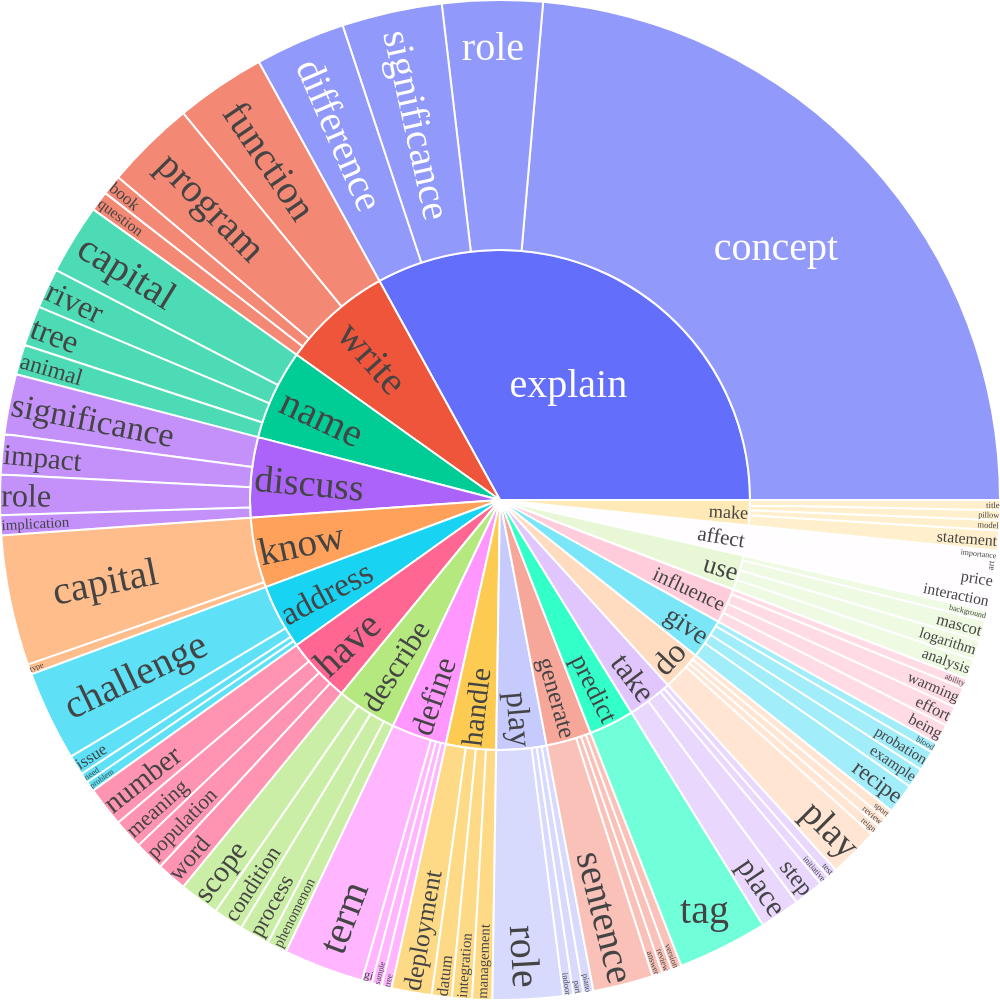}
    }
    \subfigure[ChatbotIT]{
    \includegraphics[width=.48\linewidth]{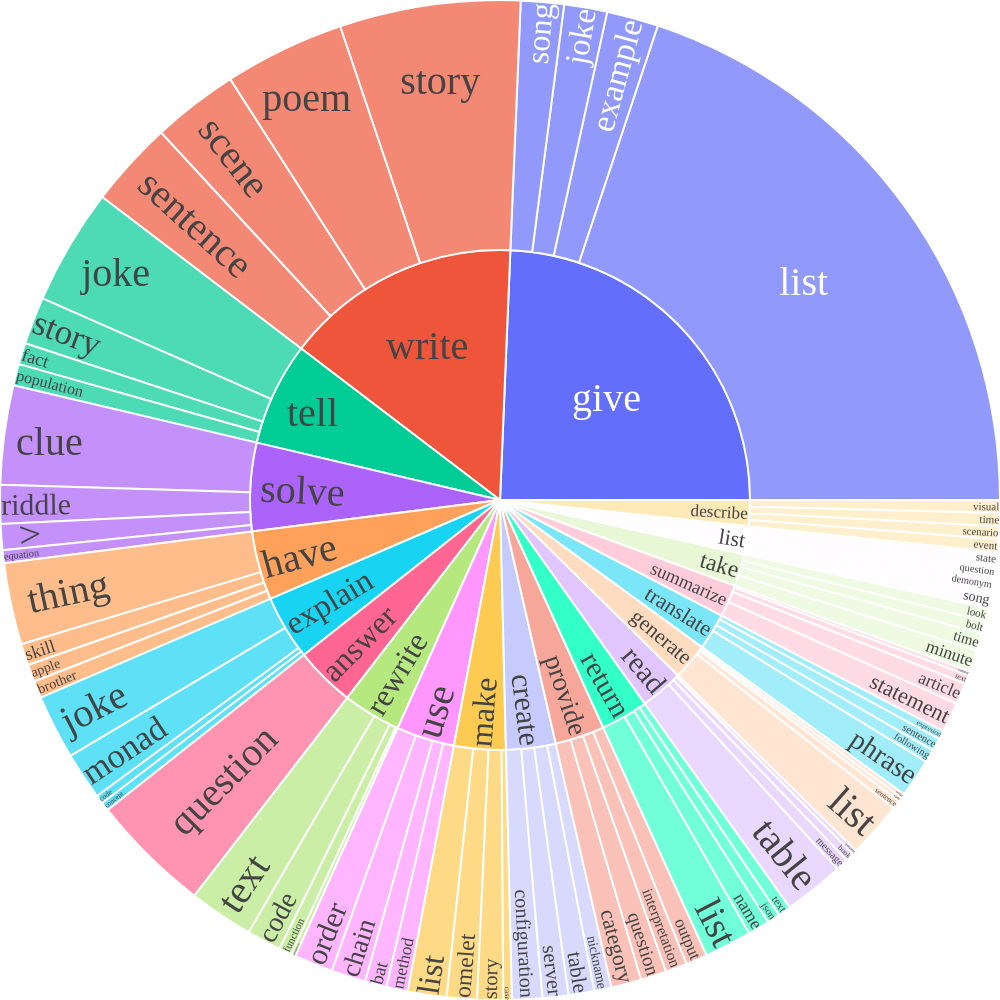}}
    \subfigure[WildChat]{
    \includegraphics[width=.48\linewidth]{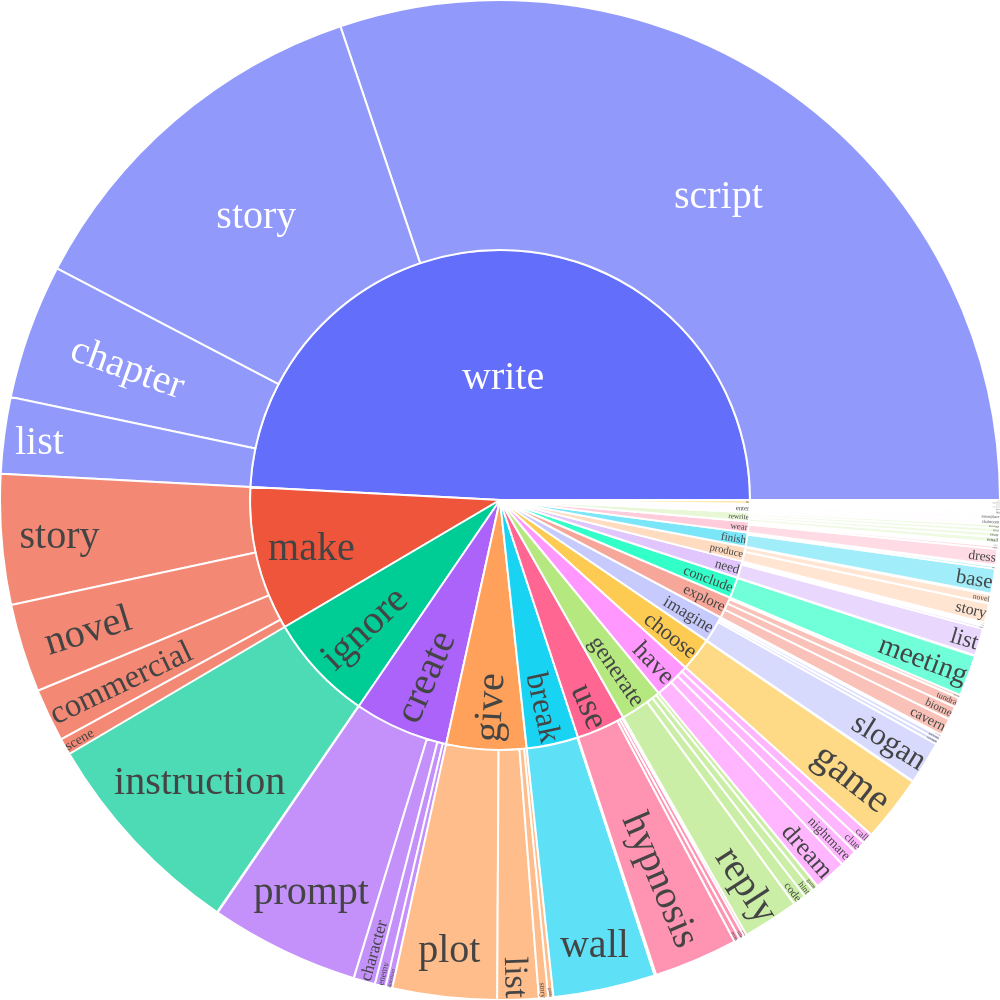}}
    \subfigure[ChatbotPA]{
    \includegraphics[width=.48\linewidth]{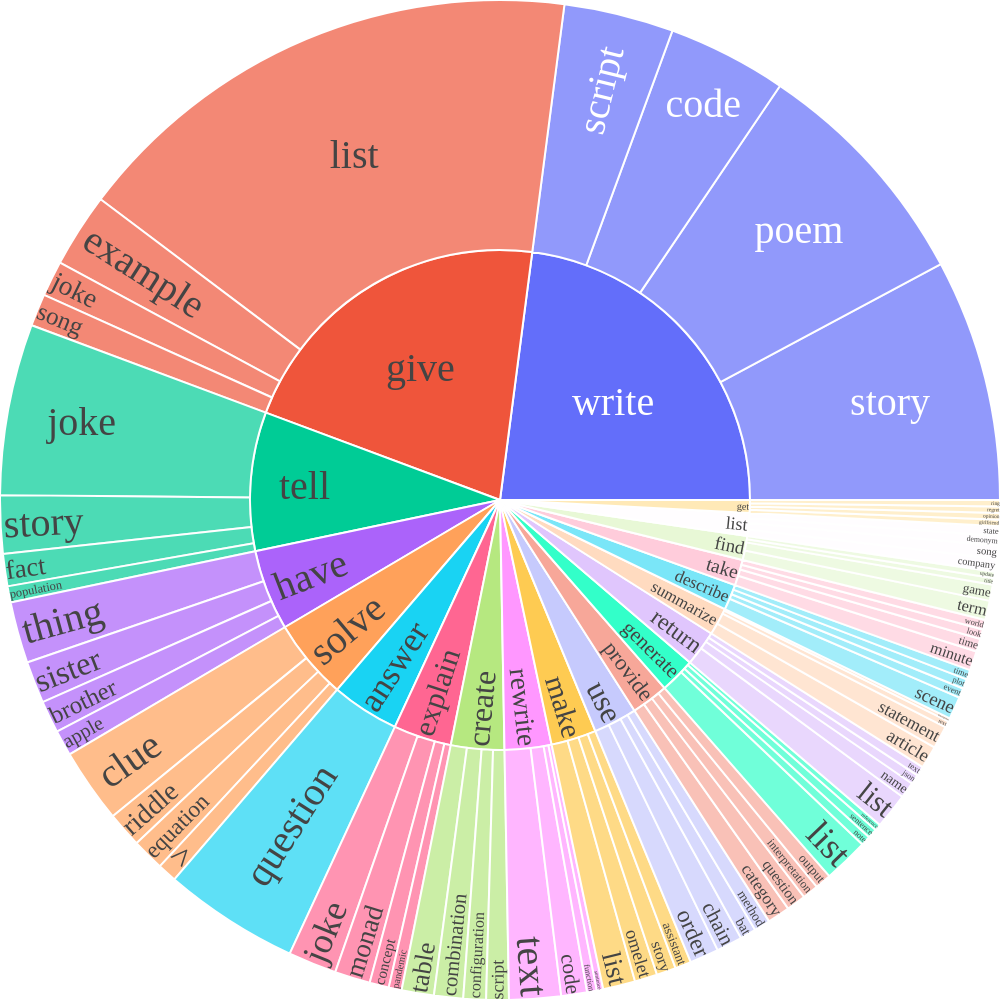}} 
    \caption{The top 20 most common root verbs (inner circle) and their top 4 direct noun objects (outer circle) in the instructions of four datasets. }
    \label{fig:pie_chart}
\end{figure}
We show the top 20 verbs and corresponding top 4 nouns of instructions of overall four datasets in Figure~\ref{fig:pie_chart}. We refer to the visualization code of Self-instruct~\cite{self-instruct}. Note that for all four datasets, we choose clients with English samples. From Figure~\ref{fig:pie_chart}, we can observe that different datasets possess diverse instruction types and distributions. For example, the top 2 verbs for Aya and WildChat datasets are (explain, write) and (write, make), respectively; While ChatbotIT and ChatbotPA, which are from the same public dataset, have a large range of keyword overlap but different quantity distributions.

\begin{figure}[H]
    \centering
    \includegraphics[width=\linewidth]{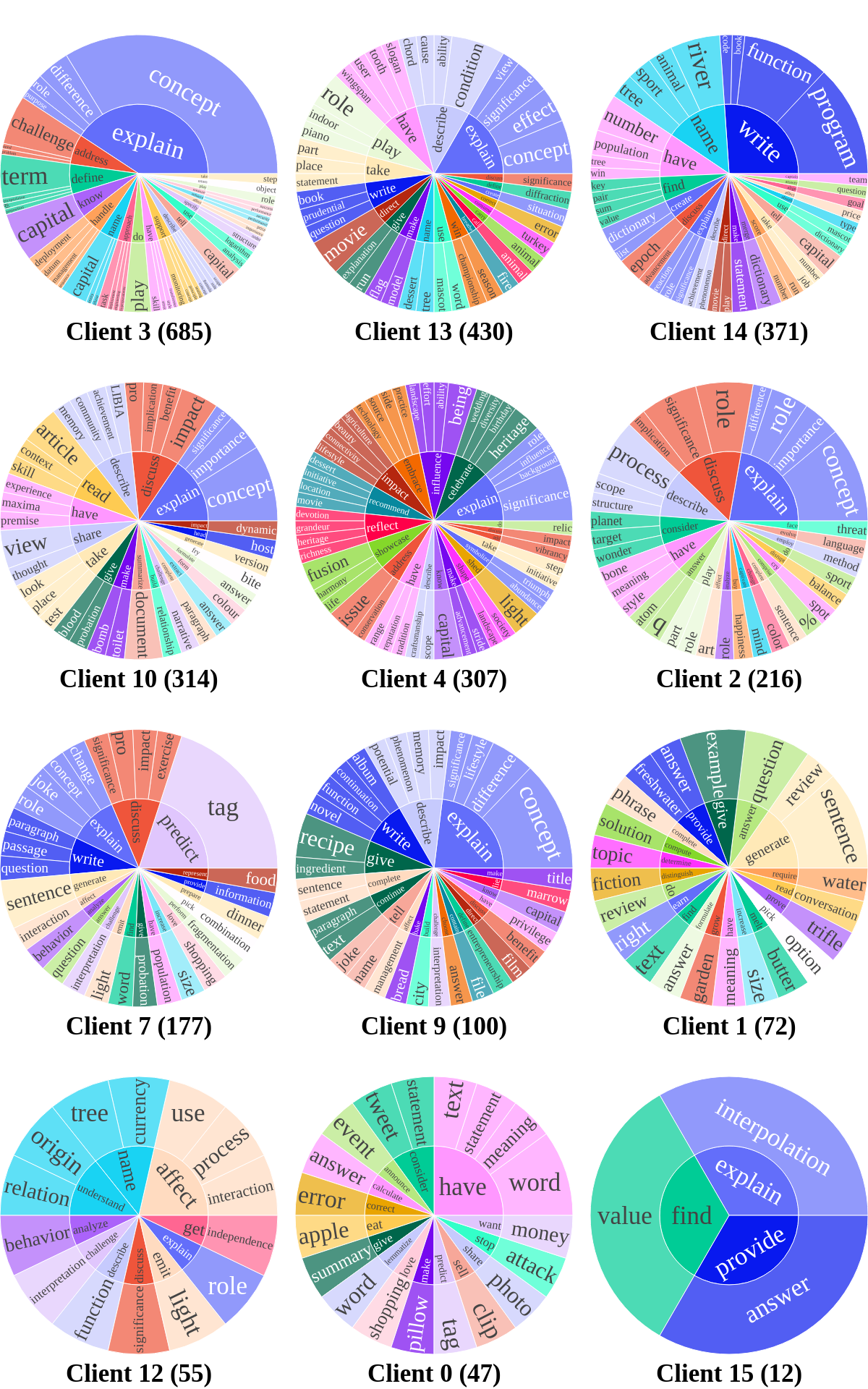}
    \caption{The top 20 most common root verbs (inner circle) and their top 4 direct noun objects (outer circle) in the instructions of \textbf{Aya} for 12 different English clients. We select English clients on purpose.}
    \label{fig:aya_big_pie}
\end{figure}

\begin{figure}[H]
    \centering
    \includegraphics[width=\linewidth]{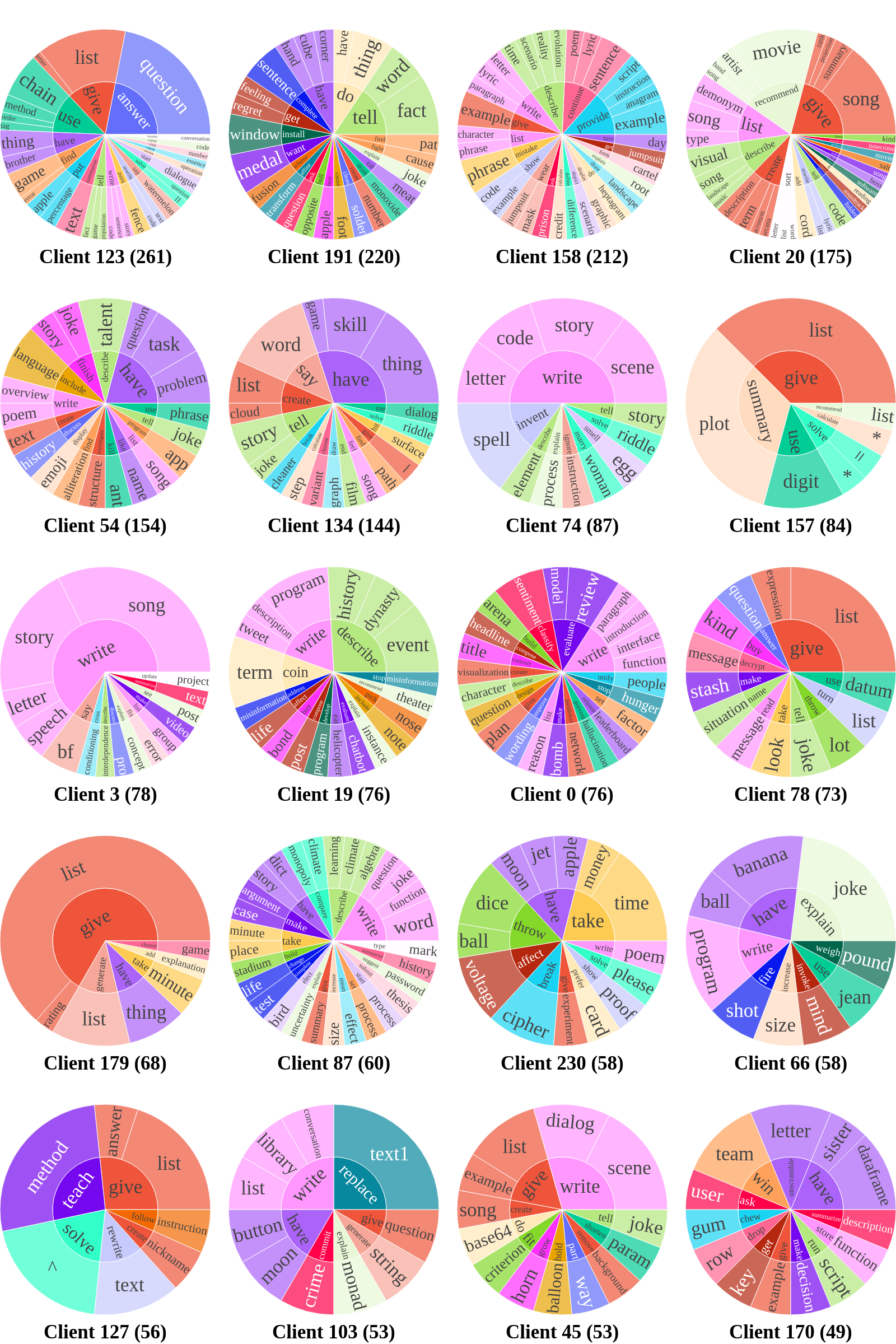}
    \caption{The top 20 most common root verbs (inner circle) and their top 4 direct noun objects (outer circle) in the instructions of \textbf{ChatbotIT} for 20 different clients.}
    \label{fig:chatbot_big_pie}
\end{figure}

\begin{figure}[H]
    \centering
    \includegraphics[width=\linewidth]{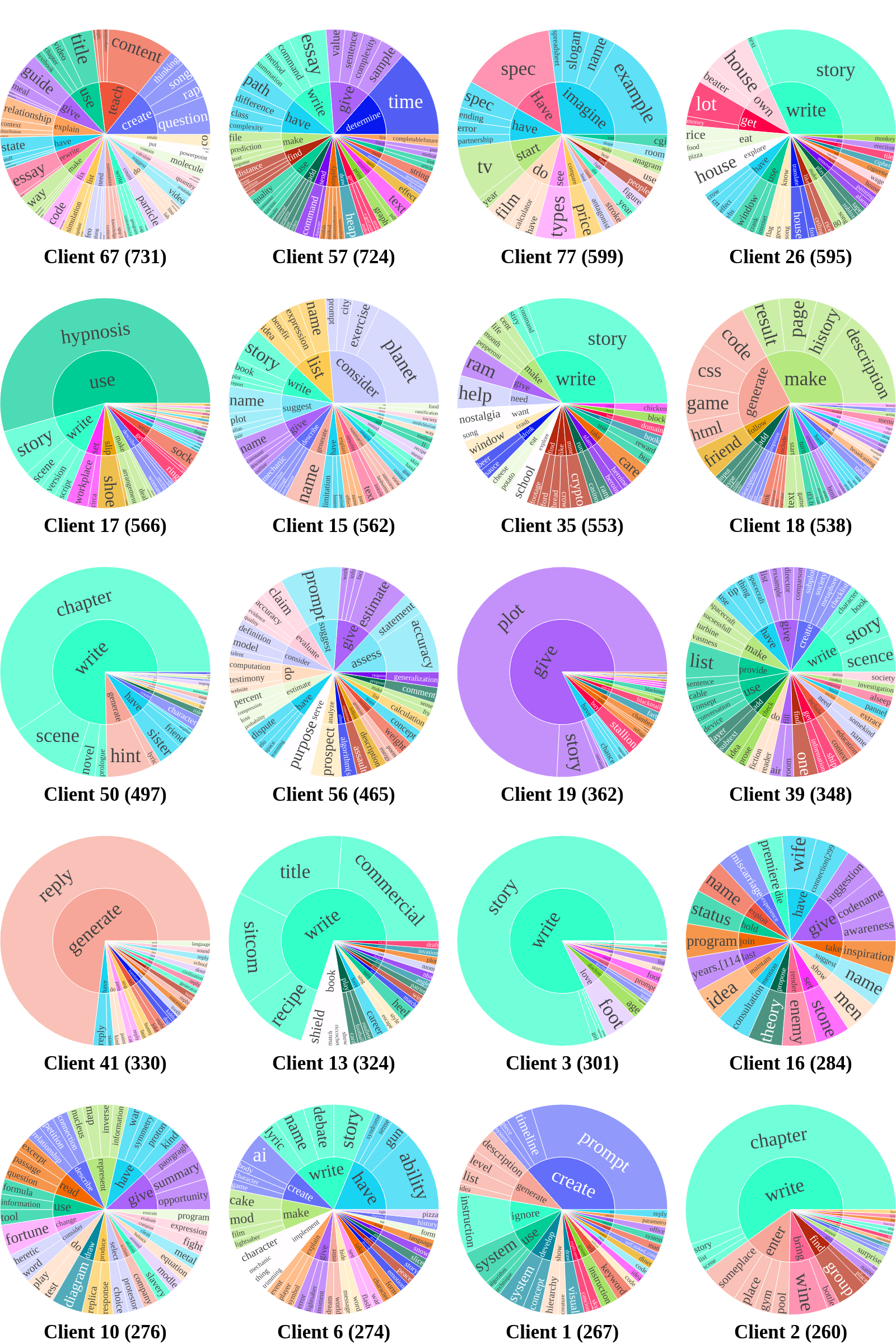}
    \caption{The top 20 most common root verbs (inner circle) and their top 4 direct noun objects (outer circle) in the instructions of \textbf{WildChat} for 20 different clients.}
    \label{fig:wildchat_big_pie}
\end{figure}

\begin{figure}[H]
    \centering
    \includegraphics[width=\linewidth]{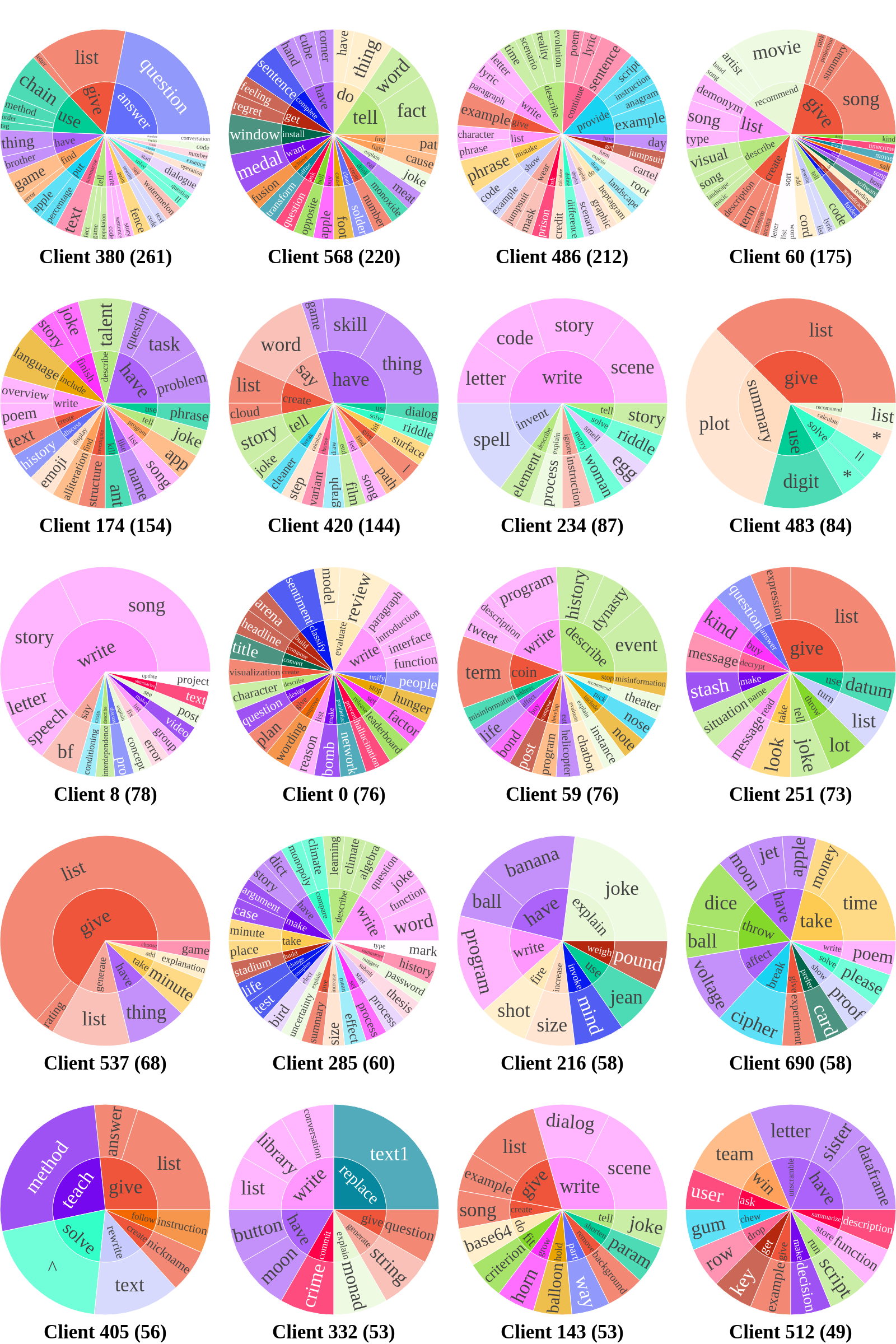}
    \caption{The top 20 most common root verbs (inner circle) and their top 4 direct noun objects (outer circle) in the instructions of \textbf{ChatbotPA} for 20 different clients.}
    \label{fig:chatbotPA_big_pie}
\end{figure}

We also show the top 20 verbs and corresponding top 4 nouns of instructions of individual clients from four datasets in Figure~\ref{fig:aya_big_pie}, Figure~\ref{fig:chatbot_big_pie}, Figure~\ref{fig:wildchat_big_pie} and Figure~\ref{fig:chatbotPA_big_pie}. For all four datasets, we choose clients with English samples to present verbs and nouns in their instructions.

\subsection{Quality evaluation}
We conduct data quality evaluation with the pre-trained Llama2-7B~\cite{llama2} and the IFD metric~\cite{li2023quantity}. IFD is a quality evaluation metric, qualifying the instruction-following difficulty of the given model as the data quality. It has been widely used in ~\cite{li2024superfiltering, zhang2024fedpit, liu2023makes}. From Figure~\ref{fig:quality_distribution}, we can see that clients in four datasets have various data qualities. This indicates these four federated datasets demonstrate quality heterogeneity, which is an inherent property of real data sets.

\subsection{t-SNE visualization}
We also implement the t-SNE instruction-response embedding in four datasets. Here the \textit{'text-embedding-ada-002'} from OpenAI is utilized as the feature extraction model. We randomly select 10 clients from each dataset and use the t-SNE two-dimensional visualization to demonstrate the data heterogeneity from each client. From Figure~\ref{fig:tsne}, we could see that data points from the same client cluster in the feature space. This is particularly evident in Fed-Aya and Fed-WildChat, demonstrating data heterogeneity within the dataset.

\subsection{Discussion about data privacy and safety}

\label{app:datasets}
The base datasets we construct ours from have already undergone screenings for safety and privacy. For example, in Chatbot-arena Conversations dataset, most conversations that contain personally identifiable information have been moved. Fed-ChatbotIT, Fed-ChatbotPA and Fed-WildChat, which are based on Chatbot-arena Conversations dataset and WildChat dataset, may contain unsafe or toxic interacts, but they are kept so that these datasets can be better used to study AI safety and simulate real-world dialogue scenarios.

\section{Experiments}

\begin{table}[t]
  \caption{Experimental setups of all datasets. `Local epochs' denotes the number of training epochs in local training. In the column of `Clients', x/y denotes that there are y clients in total and we same x clients for each round. }
  \label{tab:setup}
  \centering
  \begin{tabular}{l|c|ccc}
    \toprule
    Dataset & Local Epochs & Clients & Local Steps & Global Rounds \\
    \midrule
    Fed-Aya & 5 & 4/38 & 10 & 200 \\
    Fed-ChatbotIT  & 10 & 10/237 & 5 & 100 \\
    Fed-WildChat  & 5 & 5/100 & 10 & 100 \\
    Fed-WildChat (Multi-Turn) & 20 & 3/50 & 10 & 50 \\
    Fed-ChatbotPA  & 10 & 10/747 & Dynamic & 200 \\
    \bottomrule
  \end{tabular}
\end{table}

\subsection{Experimental setups}
\label{app:exp_setup}

We report the detailed experimental setups in Table~\ref{tab:setup}.
For each table, we randomly sample two clients to conduct local training and average their performance as the final results of `local training' in the table. We use dynamic local steps for local training in ChatbotPA. We calculate the probability of a user being selected in a round given parameters such as the total number of rounds, the number of clients sampled per round, and the total number of clients. We then adjust the local training steps for each client based on their sampling probability and data volume, ensuring that each client's data can undergo about three epochs of training in total.
Our experiments were mainlt conducted on a machine equipped with an NVIDIA GeForce RTX 3090 GPU with 24 GB of VRAM. Experiments on Fed-WildChat(Multi-Turn) were conducted on a machine equipped with an NVIDIA A40 with 48GB of VRAM. 

\subsection{Evaluation}
\begin{figure}[H]
    \centering
    \includegraphics[width=.9\linewidth]{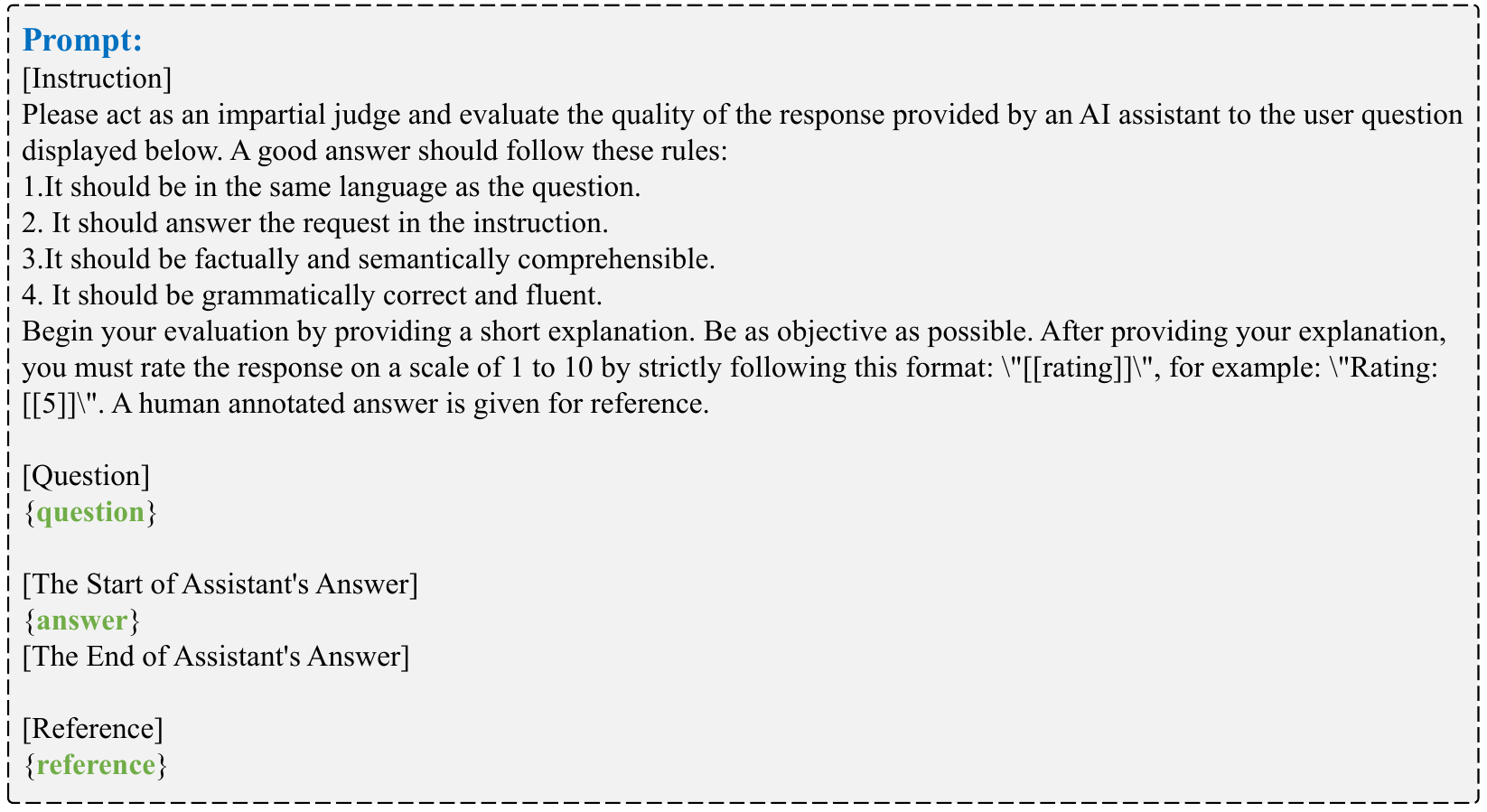}
    \caption{Prompt template used in GPT-4 judge.}
    \label{fig:ref_gpt4_prompt}
\end{figure}

Here, we show the prompt template used in GPT-4 Judge in Figure~\ref{fig:ref_gpt4_prompt}. For the specific dataset Aya, we utilize some test samples from the raw dataset, where each sample contains a question and a reference answer. We infer the tested model with the question and obtain an answer. Then we fill in the "question", "answer" and "reference" blanks of the template.

\section{More details about differential privacy}
\label{app:dp}

\begin{table}[t]
  \caption{MT-Bench on WildChat with Differential Privacy and fixed $\sigma$. FedLLM with DP($\sigma$=0.1) still outperforms local and differential privacy costs slight model performance while ensuring user-level differential privacy.}
  \label{wildchat-privacy-sigma}
  \centering
  \begin{tabular}{lllllll}
    \toprule
    local(813) &local(1702) & FedAvg & FedDP-$1e^{-3}$ & FedDP-$1e^{-2}$ & FedDP-0.1  & FedDP-1\\
    \midrule
    3.5375 & 4.0875 & 4.6875 & 4.6750 & 4.5500 & 4.5375 & 1.6875 \\
    \bottomrule
  \end{tabular}
\end{table}

\subsection{Definition}
Differential privacy (DP)~\cite{dwork2006calibrating} has emerged as a broadly recognized framework for safeguarding privacy in statistical analyses. Through DP, we can perform computations on extensive datasets while ensuring that individual data points remain indistinguishable, thereby protecting personal privacy.

In general, we use privacy parameters $\epsilon$ and $\delta$ to formally define DP.
Specifically, a randomized mechanism $M: \mathcal{D} \rightarrow \mathcal{R}$ is $(\epsilon, \delta)$-differential private for $\epsilon > 0$ and $\delta \in [0, 1)$ if for any two neighboring datasets $D, D' \in \mathcal{D}$ differing by at most one entry and for any subset of outputs $R \subseteq \mathcal{R} $ it holds that
\[
\mathbb{P}(M(D) \in R) \leq \exp(\epsilon) \mathbb{P}(M(D') \in R) + \delta.
\]

\subsection{User-level differential privacy}
\label{sec:user-level dp}
We implement user-level differential privacy(UDP)~\cite{DBLP:journals/corr/abs-2003-00229} in our experiments. Following~\cite{DBLP:journals/corr/abs-2003-00229}, we use the Gaussian mechanism that employs the $L_{2}$ norm sensitivity. It adds zero-mean Gaussian noise with variance $\sigma^2 \mathbf{I}$ to each coordinate of the function output $r(D)$ as follows:
\[
\mathcal{M}(D) = r(D) + \mathcal{N}(0, \sigma^2 \mathbf{I}),
\]
where $\mathbf{I}$ is an identity matrix of the same size as $r(D)$. The sensitivity of the function $r$ is expressed as:
\[
\Delta r = \max_{D,D' \in \mathbf{D}} \|r(D) - r(D')\|_2,
\]
which provides an upper bound on the necessary perturbation to its output for privacy preservation. By appropriately selecting the value of $\sigma$, this mechanism satisfies $(\epsilon, \delta)$-differential privacy.

\subsection{Experiment details of differential privacy}
In our experiments with UDP, we use WildChat dataset. For convenience, the batch size is 1 in all DP experiments and other settings are the same as single-turn WildChat experiment setting, see details in Section~\ref{section:experiments}.

We add Gaussian noise cautiously controlled by $\sigma$ when local clients upload their local models to the server, ensuring User-level differential privacy. Following~\cite{DBLP:journals/corr/abs-2003-00229}, the value of $\sigma$ is calculated by: 
\[
\sigma = \delta_{l} \frac{\sqrt{2qN \ln \frac{1}{\delta}}}{\epsilon}
\]
where $q$ is the sample fraction of clients each round, $N$ is the federated learning communication round, $(\epsilon, \delta)$ is the DP parameters and $delta_{l}$ is decided as follows:
\[
\delta_{l} = \frac{2 \eta C }{\frac{\left| D\right|}{\left| n \right|}}
\]
where $\eta$ is the learning rate, $C$ is the maximum of gradients, $\left| D\right|$ is the size of dataset and $\left| n\right|$ is the number of clients. Note that when expressed with $(\epsilon,\delta)$, smaller $\epsilon$ means smaller privacy budget, in other words, better privacy and usually lower performance. However, when expressed with $\sigma$, larger $\sigma$ means better privacy and usually lower performance. 

Our experiment results are shown in Figure~\ref{fig:wildchat_dp}
and Table~\ref{wildchat-privacy-sigma}. Figure~\ref{fig:wildchat_dp} shows that FedLLM with $(0.01,1e^{-4})$-DP still outperforms local training and differential privacy costs slight model performance while ensuring user-level differential privacy.

In fact, the $\sigma$ calculated in Section~\ref{sec:user-level dp} is a proper bound when ensuring user-level differential privacy with given $(\epsilon,\delta)$. We also conduct experiments with fixed $\sigma$ and results are shown in Table~\ref{wildchat-privacy-sigma}.

%%%%%% Big Images %%%%%%%%%%%
\newpage

%%%%%%%%%%%%%%%%%%%%%%%%%%%%%%%%%%%%%%%%%%%%%%%%%%%%%%%%%%%%
\vfill
\newpage

\end{document}